\documentclass{article}

\usepackage{arxiv}
\usepackage{tabularx}
\usepackage[ruled,linesnumbered,commentsnumbered]{algorithm2e}
\usepackage[utf8]{inputenc}

\usepackage[bookmarks=false]{hyperref}
\usepackage{listings}
\usepackage{color}
\usepackage{graphicx}%
\usepackage[utf8]{inputenc} 
\usepackage[T1]{fontenc}    
\usepackage{hyperref}       
\usepackage{url}            
\usepackage{booktabs}       
\usepackage{amsfonts}       
\usepackage{nicefrac}       
\usepackage{microtype}      
\usepackage{lipsum}
\usepackage{mathptmx}
\usepackage{hyperref}
\usepackage{graphicx}
\usepackage{subcaption} 
\usepackage{footnote}
\usepackage{amsmath}
\usepackage{amssymb}
\usepackage{mathrsfs}
\usepackage{array}
\usepackage{cite}
\usepackage{multirow}
\usepackage{siunitx}
\usepackage{tablefootnote}
\usepackage{mathtools}
\usepackage{commath}
\usepackage{scalerel}
\usepackage[utf8]{inputenc}
\title{A Neurochaos Learning Architecture for Genome Classification }
\author{
Harikrishnan NB, Pranay SY, Nithin Nagaraj\\
Consciousness Studies Programme,\\ National Institute of Advanced Studies,\\ Indian Institute of Science Campus, Bengaluru, India. \\  \texttt{harikrishnannb@nias.res.in, mail@pranaysy.com, nithin@nias.res.in  } \\
}
\begin{document}
\maketitle

\begin{abstract}
There has been empirical evidence of  presence of non-linearity and chaos at the level of single neurons in biological neural networks. The properties of chaotic neurons inspires us to employ them in artificial learning systems. Here, we propose a Neurochaos Learning (NL) architecture, where the neurons used to extract features from data are 1D chaotic maps. ChaosFEX+SVM, an instance of this NL architecture, is proposed as a hybrid combination of chaos and classical machine learning algorithm. We formally prove that a single layer of NL with a finite number of 1D chaotic neurons satisfies the Universal Approximation Theorem with an exact value for the number of chaotic neurons needed to approximate a discrete real valued function with finite support. This is made possible due to the  \emph{topological transitivity} property of chaos and the existence of uncountably infinite number of \emph{dense orbits} for the chosen 1D chaotic map. The chaotic neurons in NL get activated under the presence of an input stimulus (data) and output a chaotic firing trajectory. From such chaotic firing trajectories of individual neurons of NL, we extract \emph{Firing Time}, \emph{Firing Rate}, \emph{Energy} and \emph{Entropy} that constitute ChaosFEX features. These ChaosFEX features are then fed to a Support Vector Machine with linear kernel for classification. The effectiveness of chaotic feature engineering performed by NL (ChaosFEX+SVM) is demonstrated for synthetic and real world datasets in the low and high training sample regimes. Specifically, we consider the problem of classification of genome sequences of SARS-CoV-2 from other coronaviruses (SARS-CoV-1, MERS-CoV and others). With just one training sample per class for 1000 random trials of training, we report an average macro F1-score $> 0.99$ for the classification of SARS-CoV-2 from SARS-CoV-1 genome sequences. Robustness of ChaosFEX features to additive noise is also demonstrated. We foresee NL architecture being designed with novel combinations of chaotic feature engineering with other machine learning algorithms in future applications.
%
\end{abstract}

\maketitle

\section{Introduction}
Chaos emanates from the variety of behaviours exhibited by a simple deterministic non-linear dynamical system. This range of behaviours vary from periodic to random-like~\cite{devaney1993first}. The discovery of chaos theory also contributed to research in neuroscience, climate science, cryptography, epidemiology etc. Chaotic behaviours are well studied in neuroscience, especially at the level of neurons~\cite{therechaos2},~\cite{tsuda1991chaotic}. Neuronal cells exhibit a large range of firing patterns such as repetitive pulses (periodic), quasi-periodicity, and bursts of action potentials~\cite{therechaos2}. These firing patterns are driven by the external stimuli such as variations in the ionic environment driven by the effects of neuromodulators. This variability in the firing patterns indicates the presence of non-linearity and chaos at the level of neuron, axon etc. Such conclusions were inferred using classical intracellular electrophysiological recordings of action potentials in single neurons along with the help of macroscopic models~\cite{therechaos2}. 

 There has been extensive research to develop mathematical models to capture the behaviour of a biological neuron. One such model is the Hindmarsh and Rose model~\cite{hindmarsh1984model}. This model captures the oscillatory burst discharges found in real neuronal cells and is capable of exhibiting chaos. A detailed study that supports the presence of chaos in the brain at various spatiotemporal scales and mathematical neuronal models exhibiting chaos is provided in~\cite{therechaos1} and~\cite{therechaos2}. 

The current state of the art deep learning algorithms do not incorporate the rich properties of chaos at the level of artificial neurons. This research gap has been  highlighted in our earlier work~\cite{harikrishnan2019novel}, \cite{balakrishnan2019chaosnet}. The presence of chaos in biological neural networks at various spatiotemporal scales~\cite{tsuda1991chaotic} makes chaotic neurons a potential candidate to compete with current artificial neurons used in Artificial Neural Networks (ANNs).
In our earlier work, we proposed the \verb+ChaosNet+ architecture~\cite{balakrishnan2019chaosnet}, where we used chaotic 1D Generalized Lur\"{o}th Series (GLS) neurons for extracting nonlinear features from the data for solving classification tasks. Further, in~\cite{harikrishnanneurochaos}, we augment the nonlinear features extracted from a single layer of GLS neurons with a Support Vector Machine classifier (SVM) trained using a linear kernel (\verb+ChaosNet+ + SVM). The efficacy of \verb+ChaosNet+ + SVM in low training sample regime is highlighted in~\cite{harikrishnanneurochaos} for Iris dataset and synthetically generated data. 

In this work, we propose an overarching architecture titled `Neurochaos Learning' (NL) that generalizes our previous research (\verb+ChaosNet+~\cite{balakrishnan2019chaosnet}, \verb+ChaosNet+ +SVM~\cite{harikrishnanneurochaos}). We contrast NL with ANN and further provide mathematical justification of the power of chaos that is employed in NL (at the level of individual neurons) in approximating a large class of discrete nonlinear functions (real-valued with finite support) - by proving a version of the universal approximation theorem. We also highlight the advantages of chaotic feature engineering which is implicit in NL architecture for the classification of coronavirus genome sequences in the high training sample as well as low training sample regimes.

Genomics research addresses the study of the roles of multiple genetic factors and its interaction with the environment~\cite{genomics_overview}. This research was enabled by the Human Genome Project. There were two key motivations for the development of Human Genome Project: (a) to exploit the global views of genome could speed up the biological research, (b) to attack problems in a unbiased and comprehensive way~\cite{human_genome_project}. The Human Genome Project was followed by other projects like ENCODE~\cite{dunham2012encode}, FANTOM~\cite{kawai2001functional}, Roadmap Epigenomics~\cite{kundaje2015integrative}. Because of these projects, there is an abundance in genomic data. This motivates the use of Machine Learning (ML) and Deep Learning (DL) algorithms in genomics research. 

Recent research~\cite{cnn_alipanahi2015predicting},~\cite{cnn_protein_binding} shows the effectiveness of Convolutional Neural Networks (CNNs) to model the sequence specificity of protein binding. In~\cite{cnn_zhang2016deep}, a three layer CNN was used to predict the effects of non-coding variants from genome sequence. 
Similar to CNN, Recurrent Neural Networks (RNNs) is another popular DL algorithm widely used in sequence modelling. The authors in~\cite{lanchantindeep} highlight the performance of hybrid architectures on a transcription factor binding (TFBS) site classification task. The research showcases the performance of CNN-RNN hybrid architecture in comparison to standalone CNN as well as RNN. 

DL algorithms are ideally suited when the number of instances in the training data set is very high. But this need not be the case all the time. The best example for limited training data is the outbreak of COVID-19 pandemic disease. The spread rate of the highly contagious SARS-CoV-2 virus responsible for COVID-19 is very high and has spread to all countries of the world~\cite{corona_2020review_2}. Also, the genome sequence of the virus shares a 79\% match with that of the SARS-CoV-1 viral genome and a nearly 50\% match with that of the Middle East Respiratory Syndrome Coronavirus (MERS-CoV) genome~\cite{symptoms_cov_2_2020severe}. Some of the common symptoms of COVID-19 are dry cough, shortness of breath and dyspnoea, myalgia, headache and diarrhoea~\cite{symptoms_cov_2_2020severe}.
The outbreak was declared a pandemic in March 2020 and while there have been 37 million cases reported globally, there is no accepted vaccine for COVID-19. An early identification of this deadly disease and isolation of patients from the rest of the population would have facilitated effective containment of disease spread. For this, we would need novel computational methods which can uniquely identify the signatures of SARS-CoV-2 virus from limited samples (available during the early stages of the outbreak). In such a scenario, ML algorithms need to classify from fewer instances of training samples, especially during the initial few days/weeks of the outbreak. It is in such situations, we demonstrate the usefulness of NL for classification. This could, in principle, be applied for future outbreaks of novel diseases. 

The sections in this paper are arranged as follows: Section~\ref{Section_methodology} explains the proposed NL architecture, Section~\ref{Section_dataset_details} provides the information of dataset used in this research, Section~\ref{Section_experiments} highlights the experiments conducted on synthetic as well as real world data, Section~\ref{Section_conclusion} provides the scope for future work and the closing remarks.
\section{Methodology\label{Section_methodology}}
\subsection{\label{sec:GLS_Neuron} GLS-Neuron}
The chaotic neurons we consider are piece-wise linear 1D maps known as as Generalized Lur\"{o}th Series (GLS)~\cite{dajani2002ergodic}. The tent map and the binary map are commonly used GLS maps and we use the former as chaotic neurons for the proposed architecture. The tent map is mathematically represented as follows:
\subsubsection{\label{sec:GLS} Tent Map} 
\begin{figure}[!h]
	\centering
		
		\includegraphics[scale=0.25]{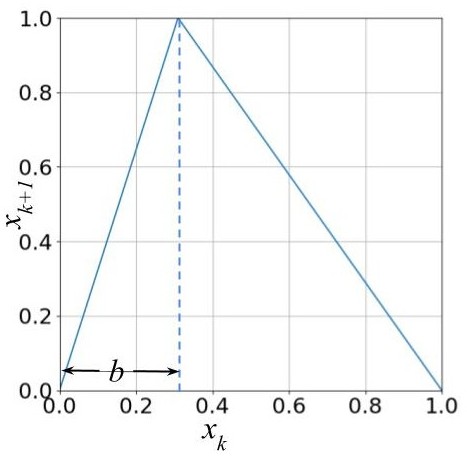}
		\caption{First return map of the GLS neuron (Skew-Tent map) used in this work~\cite{balakrishnan2019chaosnet}. }\label{fig_GLSmaps}
\end{figure}

$C_{Skew-Tent}: [0,1) \rightarrow [0,1)$ is defined as: 
\begin{eqnarray*}
C_{Skew-Tent}(x)  =  \left\{\begin{matrix}
\frac{x}{b}&, ~~~~ 0 \leq x < b, \\ 
\frac{(1-x)}{(1 - b)}&, ~~~~ b \leq x < 1, 
\end{matrix}\right. \\
\end{eqnarray*}

where $x \in [0,1)$ and $0 < b <1$. Refer to Figure~\ref{fig_GLSmaps}. 
\subsection{A Neurochaos Architecture for Learning (NL)}
The presence of neural chaos in the brain, which we describe as `neurochaos' and the earlier success of \verb+ChaosNet+ for classification tasks inspires us to propose a `Neurochaos Learning' architecture (or NL for short) in this work. The architecture consists of a multi-layer neural network built of chaotic neurons. \verb+ChaosNet+ is a particular instance of NL architecture used for classification tasks only (which makes use of a simple decision rule based on mean representation vectors)~\cite{balakrishnan2019chaosnet}. In this paper, the NL architecture includes extraction of features from the input layer followed by a linear SVM classifier. The feature extraction step using chaotic neurons is termed as ChaosFEX. Further, we can extract ChaosFEX features from a multilayer  chaotic neural network with homogeneous as well as heterogeneous chaotic neurons at different layers. These extracted ChaosFEX features can be freely combined with any of the available  classifiers or regression models from machine learning literature. Thus the proposed neurochaos learning architecture allows for a great deal of flexibility to be combined with traditional ML algorithms.
A comparison of the properties of NL with ANNs is provided in Table~\ref{Table_properties_NL_ANNs}.


\subsection{ChaosFEX}
As previously defined, ChaosFEX stands for the features extracted from the input layer of NL. In an earlier work, we have shown that such features when passed through SVM classifier with linear kernel act as an efficient representation of input data~\cite{harikrishnanneurochaos}. The chaotic features are able to provide linear separability between the classes. 

\begin{figure}
    \centerline{ \includegraphics[width=0.45\textwidth]{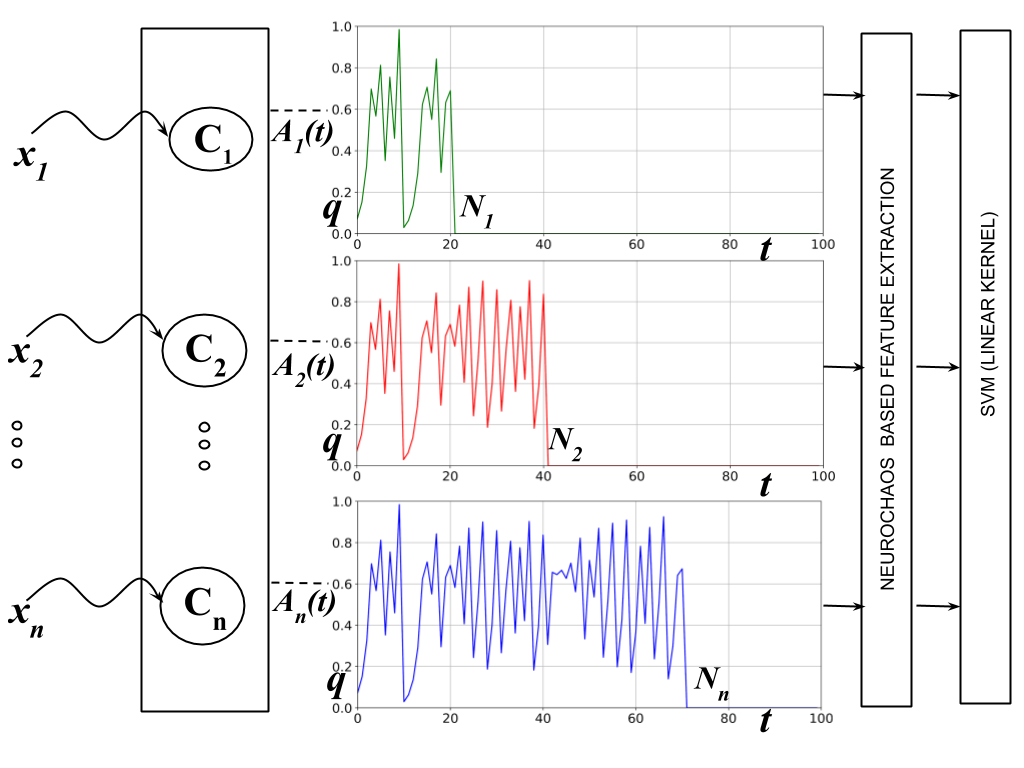}}
    
     \caption{Neurochaos Learning (NL) Architecture: 
     ChaosFEX+SVM is an instance of NL architecture. ChaosFEX extracts features from the input layer of GLS neurons ($C_1, C_2, \ldots C_n$). The stimulus or (normalized) input data to the architecture are represented as $x_1, x_2, \ldots x_n$. The chaotic neuron, say $C_k$, starts firing when it encounters the corresponding stimulus $x_k$. The trajectory of $k$-th chaotic neuron $C_k$ is represented as $A_k(t)$. The trajectory continues until it reaches the $\epsilon$ neighbourhood of the stimulus. From the chaotic trajectory $A_k(t)$, we extract firing time, firing rate, energy of the chaotic trajectory, entropy of the symbolic sequence of chaotic trajectory. These extracted features (ChaosFEX) are passed to SVM classifier with linear kernel.}
    \label{model_archi}
    \end{figure}

The proposed NL architecture using ChaosFEX is provided in Figure~\ref{model_archi}. The architecture consists of a single layer of GLS neurons ($C_1, C_2,\ldots C_n$). Features extracted from the  $n$ input GLS neurons constitute ChaosFEX which is followed by SVM classifier with linear kernel. All GLS neurons in the input layer have an initial neural activity of $q$ units. The skewness of GLS maps is controlled by the discrimination threshold ($b$). By varying $b$, the chaotic neurons can exhibit weak and strong chaos (as determined by the value of the Lyapunov exponent). 
The stimulus or input data to the proposed architecture are represented as $x_1, x_2, \ldots, x_n$ in Figure~\ref{model_archi}. The stimulus initiates the firing in chaotic neurons. The chaotic firing trajectory of the $k$-th GLS neuron, represented as $A_k(t)$, halts when the trajectory reaches the $\epsilon$ neighbourhood $I_k = (x_k - \epsilon, x_k + \epsilon)$ of the stimulus $x_k$. The time taken ($N_k$) for $A_k(t)$ to reach the $\epsilon$ neighbourhood of the stimulus ($x_k$) is defined as the \emph{Firing Time}~\cite{harikrishnan2019novel}. The chaotic firing is guaranteed to stop because of the topological transitivity~\cite{devaney1993first},~\cite{balakrishnan2019chaosnet} property of chaos.

Thus, for a single stimulus say $x_k$, the GLS neuron ($C_k$) outputs a chaotic trajectory. From this chaotic trajectory we extract the following features -- ChaosFEX:

\begin{enumerate}
    \item Firing Time.
    \item Firing Rate~\cite{balakrishnan2019chaosnet}.
    \item Energy of the chaotic trajectory~\cite{harikrishnanneurochaos}.
    \item Entropy of the symbolic sequence of the chaotic trajectory~\cite{harikrishnanneurochaos}.
\end{enumerate}

These extracted features are passed to SVM classifier with linear kernel. (Note: In the rest of this paper, ChaosFEX+SVM and ChaosFEX are used interchangeably.)

\begin{table*}
\centering
\caption{NL vs. ANN - a comparison of properties.\label{Table_properties_NL_ANNs}}

\begin{tabular}{|c|c|c|c|}
\hline
{\bf Properties} & {\bf ANN} & {\bf NL} & {\bf Remarks} \\ \hline
Neuron & \begin{tabular}[c]{@{}c@{}}Linear followed by a \\ nonlinear activation\end{tabular} & Non-linear and chaotic & \begin{tabular}[c]{@{}c@{}}Chaos allows for a rich set \\ of properties to be exploited.\end{tabular} \\ \hline
\begin{tabular}[c]{@{}c@{}}Output of a\\ Neuron\end{tabular} & Scalar & Variable length vector & \begin{tabular}[c]{@{}c@{}}Neurons in NL perform non-linear \\ computations as compared with simple \\ weighted linear addition in ANN.\end{tabular} \\ \hline
\begin{tabular}[c]{@{}c@{}}Universal \\ Approximation \\ Theorem \\ (UAT)\end{tabular} & Satisfies UAT & Satisfies UAT & \begin{tabular}[c]{@{}c@{}}NL satisfies UAT with an exact\\  specification on the number of neurons\\  needed for approximating a real-valued \\ discrete-time function with finite support.\end{tabular} \\ \hline
\begin{tabular}[c]{@{}c@{}}Activation \\ Functions\end{tabular} & Yes & No & \begin{tabular}[c]{@{}c@{}}The nonlinearity in ANN is provided by the \\ activation function which is not needed for NL.\end{tabular} \\ \hline
Backpropogation & Yes & No & \begin{tabular}[c]{@{}c@{}}Not currently used. NL could employ \\ backpropagation in the future if needed.\end{tabular} \\ \hline
\end{tabular}
\end{table*}

\subsection{Properties of Neurochaos~\label{Section:properties_of_neurochaos}}
\subsubsection{Nonlinearity}
Neurons in the brain are known to have a non-linear response and further found to exhibit chaotic behaviour~\cite{therechaos1}, ~\cite{therechaos2}. However, existing ANN architectures do not exhibit chaos at the level of neurons. In the case of NL, the neurons fire chaotically upon encountering an input sample. The notion of firing (trajectory) is absent in traditional ANNs (Table~\ref{Table_properties_NL_ANNs}).

\subsubsection{Robustness to Training Noise~\label{Sub_sec:robustness_to_traininig}}
To evaluate the robustness of ChaosFEX,  we conduct two sets of binary classification experiments -- (a) we train the NL using a noisy train data and test on a noiseless test data; (b) using the same hyperparameters used in experiment (a), we evaluate the performance of NL on a noiseless train and test data. The above two experiments are also evaluated for SVM with RBF kernel for a comparative performance analysis. We use two sets of simulated datasets for these experiments -- the concentric circle data (CCD) and overlapping concentric circle data (OCCD).

The governing equations for OCCD are as follows:
\begin{eqnarray}
f_1 &=& r_i  \cos(\theta) + \alpha \eta, \label{eqn_ccd_1}  \\
 f_2 &=& r_i  \sin(\theta) + \alpha \eta, \label{eqn_ccd_2} 
\end{eqnarray}
where $i = \{0,1\}$ ($i = 0$ represents Class-0 and $i = 1$ represents Class-1), $r_0 = 0.6$, $r_1 = 0.4$, $\theta$ is varied from 0 to \ang{360}, $\alpha = 0.1$,  $\eta \sim  \mathcal{N}(\mu,\,\sigma)$, normal distribution, with $\mu = 0$ and 
$\sigma = 1$. In the case of CCD, the value of $\alpha$ used in equation~\ref{eqn_ccd_1} and equation~\ref{eqn_ccd_2} is set to $0.01$. 

\begin{figure}[h!]
\centering
	\begin{subfigure}{0.45\linewidth}
		\centering
		\includegraphics[width=1\linewidth]{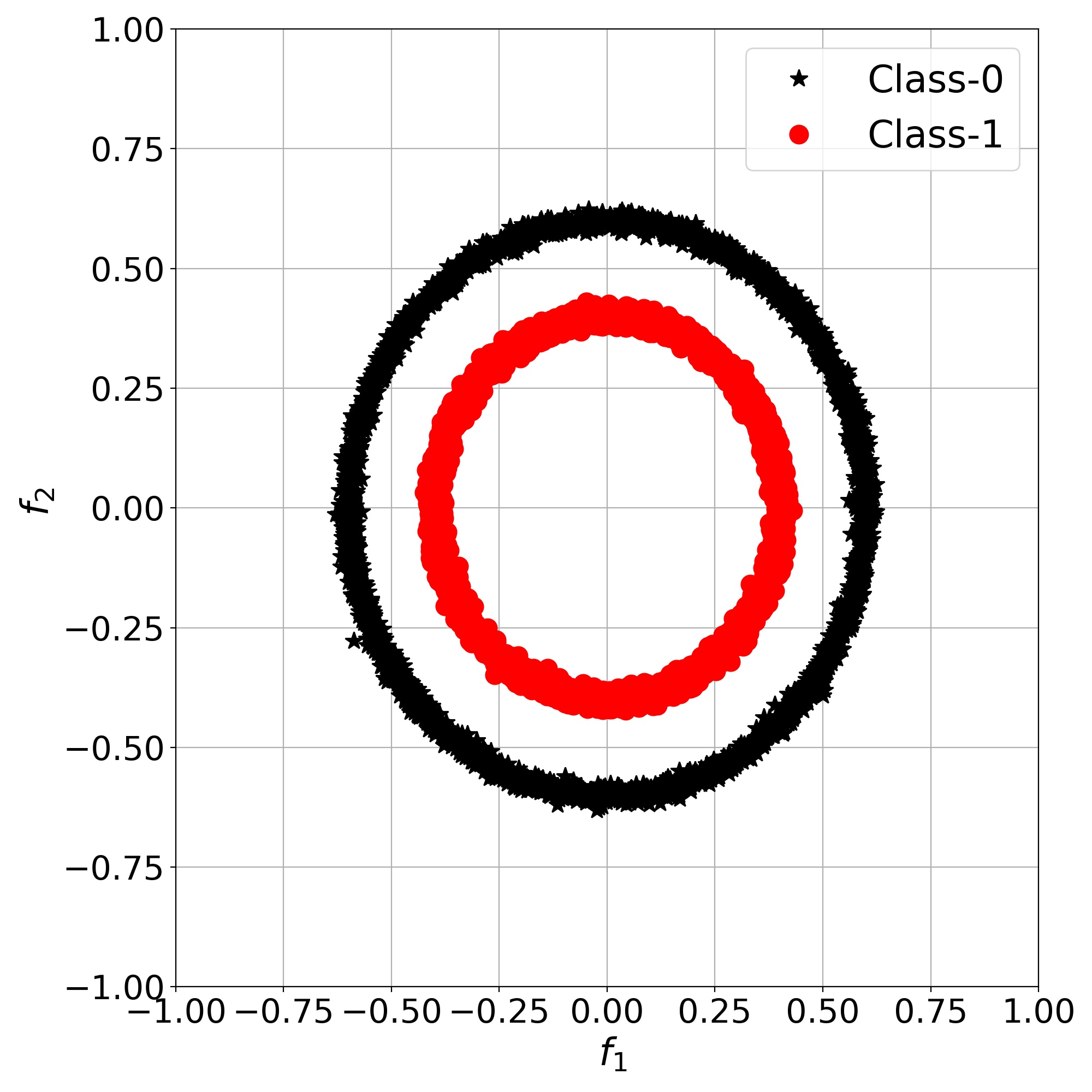}
		\caption{}\label{Fig_ccd_data}
	\end{subfigure}
	\begin{subfigure}{0.45\linewidth}
		\centering
		\includegraphics[width=1\linewidth]{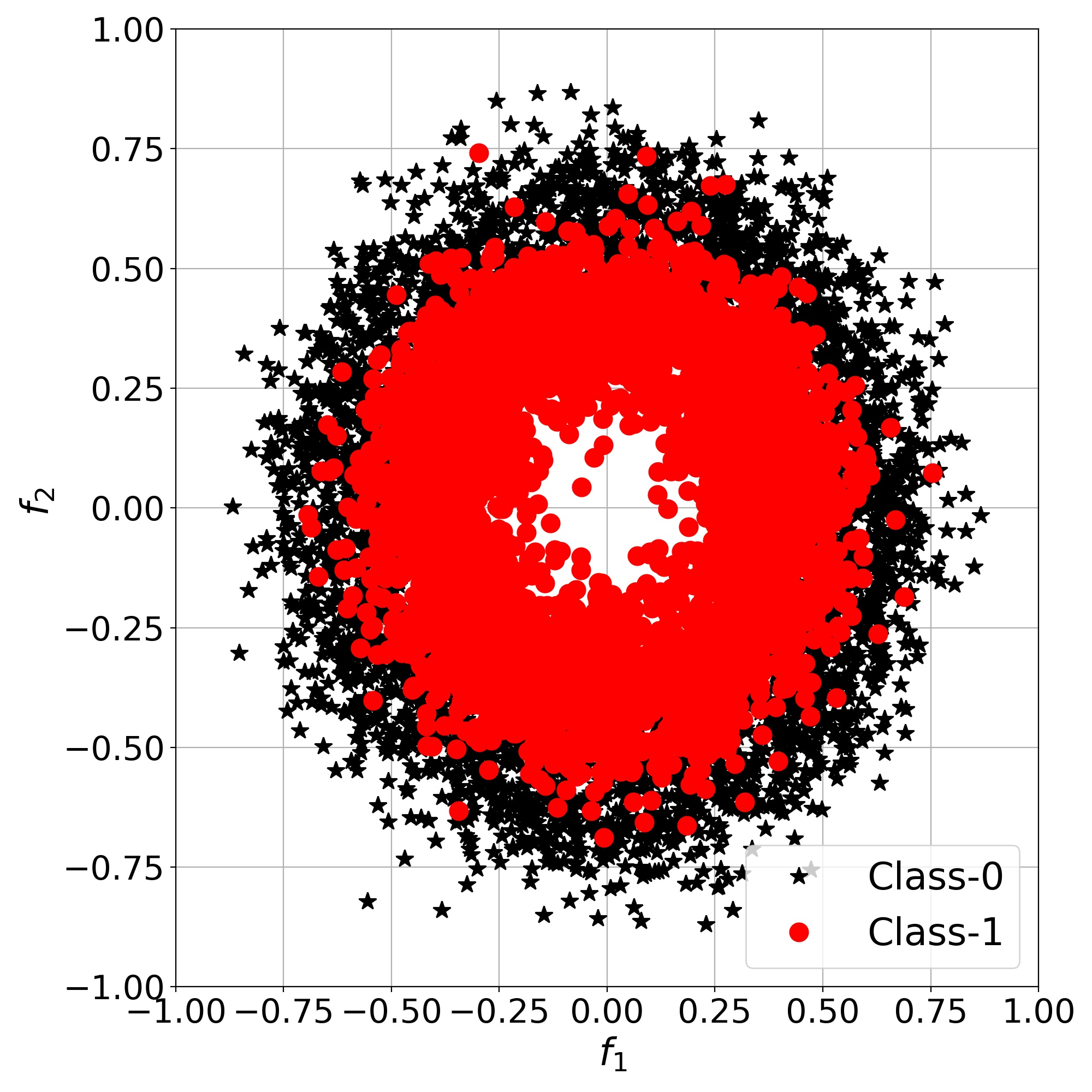}
		\caption{}\label{Fig_occd_data}
	\end{subfigure}
\caption{(\subref{Fig_ccd_data}) Concentric Circle Data (CCD).   (\subref{Fig_occd_data}) Overlapping Concentric Circle (OCCD).}

	\label{fig:3:2}
\end{figure}

Figure~\ref{Fig_ccd_data} and Figure~\ref{Fig_occd_data} represents the noiseless (CCD) and noisy (OCCD) data respectively. Table~\ref{Table_noise_training_statistics} and Table~\ref{Table_noiseless_statistics} represents the train and test data statistics used in the noisy and noiseless experiments. Table~\ref{Table_noise_experiments_results} represents the results corresponding to noise experiments.
\begin{table}[!h]
\centering
\caption{Dataset details for Expt-1 and Expt-3: Training with OCCD and Testing with CCD.}
\begin{tabular}{|c|c|}
\hline
Dataset & \begin{tabular}[c]{@{}c@{}}CCD \&\\ OCCD\end{tabular} \\ \hline
\# Classes & 2 \\ \hline
\begin{tabular}[c]{@{}c@{}}\# OCCD Training \\ instances per class\end{tabular} & (2513, 2527) \\ \hline
\begin{tabular}[c]{@{}c@{}}\# CCD Testing\\ instances per class\end{tabular} & (2513, 2527) \\ \hline
\end{tabular}
\label{Table_noise_training_statistics}
\end{table}

\begin{table}[!h]
\centering
\caption{Dataset details for Expt-2 and Expt-4: Training with CCD and Testing with CCD.}
\begin{tabular}{|c|c|}
\hline
Dataset & \begin{tabular}[c]{@{}c@{}}CCD \\ \end{tabular} \\ \hline
\# Classes & 2 \\ \hline
\begin{tabular}[c]{@{}c@{}}\# CCD Training \\ instances per class\end{tabular} & (2513, 2527) \\ \hline
\begin{tabular}[c]{@{}c@{}}\# CCD Testing\\ instances per class\end{tabular} & (1087, 1073) \\ \hline
\end{tabular}
\label{Table_noiseless_statistics}
\end{table}

\begin{table}[!h]
\centering
\caption{Test results for experiments 1 to 4.}
\begin{tabular}{|c|c|c|}
\hline
Experiment & Method & \begin{tabular}[c]{@{}c@{}}Macro F1-score \\ \end{tabular} \\ \hline
1 & \begin{tabular}[c]{@{}c@{}}Noisy\\ ChaosFEX\end{tabular} & 0.98 \\ \hline
2 & \begin{tabular}[c]{@{}c@{}}Noiseless\\ ChaosFEX\end{tabular} & 0.99 \\ \hline
3 & \begin{tabular}[c]{@{}c@{}}Noisy\\ SVM+RBF\end{tabular} & 0.67 \\ \hline
4 & \begin{tabular}[c]{@{}c@{}}Noiseless\\ SVM+RBF\end{tabular} & 0.80 \\ \hline

\end{tabular}
\label{Table_noise_experiments_results}
\end{table}
\textbf{Expt-1: Training using OCCD and Testing using CCD using NL}
The first step is to find the hyperparameters - initial neural activity ($q$), discrimination threshold ($b$) and epsilon ($\epsilon$) suitable for ChaosFEX to learn the distribution of a noiseless data from the distribution of a noisy data.
We used a small set of CCD as a validation set and found the suitable hyperparameters that  gives best performance for the validation data. The following hyperparameters, $q = 0.34$, $b = 0.499$, and $\epsilon = 0.18$ gave a best macro F1-score of  0.99. The number of class-0 and class-1 data instances belonging to the validation set are 1087 and 1073 respectively. We then retrained the NL with the hyperparameters using OCCD and tested on unseen CCD. For the unseen CCD test data we get an macro F1-score of 0.98 (Table~\ref{Table_noise_experiments_results}).

\textbf{Expt-2: Training using CCD and Testing using CCD using NL}

In this scenario, we train the NL using CCD and test using data drawn from the same distribution. In order to evaluate the robustness of ChaosFEX features, we use the same hyperparameters ($q = 0.34$, $b = 0.499$ and $\epsilon = 0.18$) used in experiment 1. For the unseen test data, we get a F1-score of 0.99.
This shows that the hyperparameters are invariant to the noise in the training data. Since the same hyperparameters work in both experiment 1 and 2, this shows the robustness of ChaosFEX to noise in training data. We could have instead chosen to train NL independently for experiment 1 and 2 in which case we would have got different set of hyperparameters which are optimum in those cases. For example, with the hyperparameters ($q = 0.22$, $b =0.96$, $\epsilon = 0.018$), we get 100\% classification accuracy for experiment 2. ChaosFEX features thus provide us with the flexibility to choose between robust and optimum hyperparameters based on the desired application. 

\textbf{Expt-3: Training using OCCD and Testing using CCD using SVM+RBF}

For SVM with RBF kernel we did a hyperparameter tuning. We choose the following values for $C$: [1.e-02, 1.e-01, 1.e+00, 1.e+01, 1.e+02, 1.e+03] and gamma : [1.e-09, 1.e-08, 1.e-07, 1.e-06, 1.e-05, 1.e-04, 1.e-03, 1.e-02,  1.e-01, 1.e+00, 1.e+01, 1.e+02, 1.e+03]. A maximum macro F1-score of 0.665 was obtained for $C = 1$ and $gamma = 0.1$. We then retrained SVM with RBF kernel with $C = 1$ and $gamma = 0.1$ using OCCD and tested on unseen CCD. For the test data we get a macro F1-score  = 0.67 (Table~\ref{Table_noise_experiments_results}).

\textbf{Expt-4: Training using CCD and Testing using CCD using SVM+RBF}

In this case, we train and test SVM with RBF kernel using CCD. For the same hyperparameters as used in experiment 3 ($C=1$, $gamma = 0.1$), we get a test macro F1-score of $0.80$. (Note: for $C=1$ and $gamma =$ `scale', we get a test accuracy of 100\%). Thus, there is a higher degradation of performance metrics (F1-score) in using the same hyperparameters for experiment 3 and 4 (0.67 and 0.80) as compared to ChaosFEX (0.98 and 0.99). Clearly, ChaosFEX features are more robust to noise.

\subsubsection{Topological Transitivity and existence of a dense orbit}
\emph{Definition:} A dynamical system ($\Sigma$) is transitive if for each two points $x, y \in \Sigma$ and for $\epsilon > 0$, there exist a $z \in \Sigma$ such that on finite number of iterations, $z$ reaches the $\epsilon$ neighbourhood of $x$ and $y$.

\emph{Definition:} A set $Y$ is a dense subset of $X$ if, for any point $x \in X$, there is a point $y$ in the subset $Y$ arbitrarily close to $x$~\cite{devaney1993first}.

We use the Topological Transitivity property of chaos and the existence of a dense orbit to prove the Universal Approximation Theorem (UAT) for GLS neurons\footnote{An example of a dense orbit on the binary map is the real number that has a binary expansion given by `$0.0~1~00~01~10~11~000~001$.\ldots'\cite{devaney1993first}. The skew-tent map can be proved to have a dense orbit by the fact that there exists a conjugacy with the binary map.}. 
%
%
\subsubsection{Universal Approximation Theorem for NL}



%
Let $f(n)$ be a discrete time real valued function having a finite support $L$. The Neurochaos Learning architecture (NL) consisting of a single layer with $L$ chaotic neurons can approximate\footnote{For quantifying this approximation, we use the sum of absolute differences as the distance metric. In other words, for any two real-valued vectors $V, W \in \mathbb{R}^m$,  $d(V,W) = \sum_{i = 1}^{m}|V_i - W_i|$.} $f(n)$. Assuming that we use a chaotic 1D map $C_i$ for the $i$-th neuron in NL, and given any desired error $\epsilon >0$, we have: 
\begin{equation}
    d(f, C) = \sum_{i = 1}^{L}|f(i) - C_i^{N_i}(q)| < \epsilon,
\end{equation}
%
%
where $q$ is the initial neural activity for all the neurons in NL, $N_i$ is the firing time of the $i$-th chaotic neuron and $C_i$ is the 1D chaotic map with the chaotic trajectory starting from $q$ a dense orbit.

\emph{Proof}: (by construction). Design an NL with one layer with exactly $L$ chaotic neurons. Let each of the neurons be initialized with $q$ and let the input to this NL be the $L$ real-valued samples of the function $f(n)$ which act as stimuli for the corresponding $L$ chaotic neurons. 

Now, for a given $\epsilon > 0$, we can always construct a neighbourhood of stimulus $I_k = (f(k) - \eta, f(k) + \eta)$, $0 < \eta < \frac{\epsilon}{2L}$ such that $C_k^{N_k}(q) \in I_k$ for the $k$-th chaotic neuron of NL. This is always possible because of the topological transitivity property of chaos defined in Section~\ref{Section_methodology} and since the chaotic trajectory starting from initial value $q$ is dense. The topological transitivity property guarantees the chaotic firing to reach the $\eta$ neighbourhood ($I_k$) of stimulus in finite number of iterations ($N_k$) for the dense orbit starting from $q$. For any given $\epsilon$, the following is true:
\begin{eqnarray}
d(f,C) &=& \sum_{i=1}^{L}
d(f(i),C_i^{N_i}),\\
&=& \sum_{i=1}^{L} \lvert f(i) - C_i^{N_i}(q) \rvert, \\
&<& \sum_{i=1}^{L} 2\eta,\\
&<& L(2\eta),\\
&<& L(2\frac{\epsilon}{2L}),\\
&=& \epsilon. 
\end{eqnarray}
Eq.(7) is true since $C_i^{N_i}(q) \in (f(i) - \eta, f(i) + \eta)$ because $N_i$ is the firing time for $C_i$ and the orbit is dense. Hence, the set of chaotic neurons $\{ C_i \}$ that constitute the input layer of NL can always approximate the function $f(n)$ with an $\epsilon$ error bound. This theorem holds true for NL constructed with chaotic neurons that satisfies the topological transitivity property and has a dense orbit. Further, having a single dense orbit implies uncountably infinite number of dense orbits. In other words, let $q$ be the initial value corresponding to a dense orbit. We can do a back iteration on $q$ infinite number of times to get a unbounded binary tree whose cardinality of paths is uncountably infinite. Any initial value chosen from the node of this unbounded binary tree will reach $q$ after a finite number of iterations. Since the trajectory starting with $q$ is a dense orbit (by assumption), hence the trajectory corresponding to any value in the node of this binary tree is also dense. Furthermore, all these are distinct dense orbits.

An advantage of the above constructive proof is that we are guaranteed to satisfy UAT for discrete time functions of finite support ($L$) with a Neurochaos Learning architecture (NL) comprising of only one input layer with  exactly $L$ chaotic neurons. Contrast this with ANNs where it is not known what is the lower bound of the number of neurons needed for UAT.
\subsection{Chaos for feature engineering}
ChaosFEX feature extraction maps the data into a high dimensional space. The input data matrix with $m$ data instances and $n$ points per data instance ($m \times n$) is mapped to $m \times 4n$. We extract $4$ features from the chaotic trajectory. In general, the input data matrix with size $m \times n$ is mapped to chaotic feature space where the new matrix dimension is $m \times kn$, where $k$ represents the number of distinct features extracted from the chaotic firing of the neurons in the input layer of NL. In \verb+ChaosNet+~\cite{balakrishnan2019chaosnet}, $k = 1$ because only a single feature (Firing Rate) was extracted. Thus, chaos based feature engineering can be incorporated in either NL (as we have done in this work) or in conventional ANNs.

\subsection{Neuromorphic Computing}
The aim of Neuromorphic computing~\cite{smith1998neuromorphic} research is to develop hardware and software architectures inspired from the structural and functional mechanisms of the brain. This involves exploiting the rich properties of biological neural networks such as robust learning, plasticity~\cite{neural_plasiticity_burk_2006} and low computation power. 
NL allows to integrate neuronal models inspired from the biological neurons. The neurons in NL can be replaced by the Hindmarsh and Rose neuronal model~\cite{hindmarsh1984model} or 
adaptive exponential integrate-and-fire model~\cite{brette2005adaptive} or other neuronal models that simulate the firing patterns of biological neurons. Such a combination of biological neuronal models and classical machine learning techniques directly contributes to Neuromorphic computing paradigms by aiding interpretability of the learning process.
\subsection{Other properties}
Deterministic chaos provides a rich variety of behaviours - periodic, quasi-periodic, eventually periodic and non-periodic orbits or trajectories, multiple co-existing attractors, fractal boundaries, various types of synchronization - all of which are being exploited in engineering applications. This adaptability of chaotic systems allows applications in several fields such as chaotic computing~\cite{kia2014chaotic_computing}, lossless compression and memory encoding~\cite{kathpalia2019novel, memory2017chaos}, chaos based cryptography~\cite{kocarev2001chaos}, chaotic neural multiplexing~\cite{nagaraj2016neural}, chaotic neural networks~\cite{aihara1990chaotic}, fractal image compression~\cite{fisher2012fractal} etc. Despite having a positive Lyapunov exponent, two chaotic systems still synchronizes under specific conditions~\cite{pecora1990synchronization}. 

The GLS neurons used in NL is proven to be useful in memory encoding~\cite{kathpalia2019novel}, lossless compression~\cite{nagaraj2009arithmetic}, cryptography~\cite{wong2010simultaneous},~\cite{nagaraj2008novel} and error control coding~\cite{nagaraj2019cantor}.  These rich properties of chaotic neurons employed in NL (possibly in conjunction with traditional ML architectures) makes it a very good competitor for ANNs.
\section{Dataset Details\label{Section_dataset_details}}
This section provides a detailed description of datasets used to evaluate the efficacy of ChaosFEX. We used synthetically generated as well as real wold datasets for our analysis. The real world dataset consists of genome sequences of SARS-CoV-2 and other coronaviruses. The synthetic data helps to visualize the non linear features extracted from ChaosFEX. 
\subsection{Synthetically Generated: Overlapping Concentric Circle Data (OCCD)}
We used synthetically generated OCCD data described in Subsection~\ref{Sub_sec:robustness_to_traininig} of Section~\ref{Section:properties_of_neurochaos} for evaluating the efficacy of ChaosFEX.
The train test distribution is provided in Table~\ref{Table_OCCD_train_test_split}. In the training set, there is a 32.8\% overlap of class-0 data instances with class-1 data instances. This overlap makes the classification task challenging. 
\begin{table}[!h]
\centering
\caption{OCCD: Number of training and testing samples used in this study.}
\begin{tabular}{|c|c|}
\hline
Dataset & \begin{tabular}[c]{@{}c@{}}OCCD \\ \end{tabular} \\ \hline
\# Classes & 2 \\ \hline
\begin{tabular}[c]{@{}c@{}}\# OCCD Training \\ instances per class\end{tabular} & (2513, 2527) \\ \hline
\begin{tabular}[c]{@{}c@{}}\# OCCD Testing\\ instances per class\end{tabular} & (1087, 1073) \\ \hline
\end{tabular}
\label{Table_OCCD_train_test_split}
\end{table}
\subsection{Classification of coronaviruses}
\subsubsection{Multiclass Classification}
The classification of SARS-CoV-2 from other coronaviruses is challenging because of its similarity across coronavirus family. We used the data provided by the authors of the paper titled ``Classification and Specific Primer Design for Accurate Detection of SARS-CoV-2 Using Deep Learning"~\cite{cnn2020accurate},~\cite{who_primer_2020}. The authors extracted data from the 2019 Novel Coronavirus Resource repository (2019nCoVR)\footnote{\url{https://bigd.big.ac.cn/ncov/?lang=en}}. All available sequence under the query Neuclotide Completeness = ``\emph{Complete}" AND host = ``\emph{homo sapiens}" were downloaded by~\cite{cnn2020accurate}. After removing the recurring sequences, a 533 unique sequences of different length (1,260 - 31,029) base pair sequences were used for conducting experiments. The data details is provided in Table~\ref{Table_multi_class}. A five class classification problem is formulated with this dataset. The reasoning behind grouping of viruses into distinct classes is provided in~\cite{cnn2020accurate}.
\begin{table}[!h]
\centering
\caption{\label{Table_multi_class}Multiclass Classification: Total data instances per class for the five class classification problem.}
\begin{tabular}{|c|c|}
\hline
Data      & No. of samples \\ \hline
Class-0 & 66         \\ \hline
Class-1  & 240            \\ \hline
Class-2  & 162            \\ \hline
Class-3  & 75            \\ \hline
Class-4 & 10              \\ \hline
\end{tabular}
\end{table}

\begin{itemize}
    \item Class-0: SARS-CoV-2 (66 samples).
\item Class-1: MERS-CoV (240 samples).
\item Class-2: HCoV-OC43 (132 samples), HCoV-229E (22 samples), HCoV-4408 (2 samples), HCoV-EMC (6 samples) .
\item Class-3: HCoV-NL63 (58 samples), HCoV-HKU1 (17 samples).
\item Class-4: SARS-CoV (7 samples), SARS-CoV P2 (1 sample), SARS-CoV HKU-39849 (1 sample), SARS-CoV GDH-BJH01 (1 sample).
\end{itemize}
\subsubsection{Binary Classification (SARS-CoV-2 vs. Others)}
ORFlab gene consists of two-thirds in Coronaviruses' genome sequence. This signifies the importance of classifying SARS-CoV-2 from similar viruses. In~\cite{cnn2020accurate}, the authors downloaded the genome sequence corresponding to the following query: gene= ``ORF1ab" AND host= ``homo sapiens" AND ``complete genome"~\cite{cnn2020accurate}. The recurring sequences are removed and 45 data instances corresponding to SARS-CoV-2 are grouped as class-0 and the remaining data instances belonging to other viruses are grouped as class-1. Table~\ref{Table_binary_class} represents the data instances per class for binary classification problem.
\begin{table}[!h]
\centering
\caption{\label{Table_binary_class}SARS-CoV-2 vs. Others: Binary Classification data instances per class.}
\begin{tabular}{|c|c|}
\hline
Data      & No. of samples \\ \hline
Class-0 & 45         \\ \hline
Class-1  & 339            \\ \hline
\end{tabular}
\end{table}
\begin{itemize}
\item Class-0: SARS-CoV-2 (45 samples).
\item Class-1: MERS-CoV (180 samples), HCoV-OC43 (105 samples), HCoV-NL63 (29 samples), HCoV-HKU1 (13 samples), HCoV-4408 (2 samples), HCoV-229E (3 samples), HCoV-EMC ( 3 samples),  HAstV-VA1 (1 sample),  HAstV-BF34 (1 sample), HMO-A (1 sample), HAstV-SG (1 sample).
\end{itemize}

\subsubsection{SARS-CoV-2 vs. SARS-CoV-1}
For the binary classification of SARS-CoV-2 genomes from SARS-CoV-1 genomes, a total of 4498 and 101 genome sequences respectively were obtained from multiple data repositories until early April 2020. 3930 SARS-CoV-2 sequences were obtained from GISAID, 407 from GenBank, and the remaining from Genome Warehouse, CNGBdb and NMDC databases through the China National Center for Bioinformation~\cite{zhao20202019}. All SARS-CoV-1 sequences were obtained from GenBank. All sequences were chosen with the filters Nucleotide Completeness = ``Complete" AND host = ``homo sapiens". Accession IDs for all sequences as well as acknowledgement as provided in the GitHub repository~\footnote{\url{https://github.com/HarikrishnanNB/genome_classification/tree/master/sequence_usage_acknowledgements}}. This additional data is not used in~\cite{cnn2020accurate}. The data instance per class is provided in Table~\ref{Table_SARS-CoV-2_vs_SARS-CoV-1} (Class-0: SARS-CoV-2 and Class-1: SARS-CoV-1) .
\begin{table}[!h]
\centering
\caption{\label{Table_SARS-CoV-2_vs_SARS-CoV-1}SARS-CoV-2 vs. SARS-CoV-1: Total number of data instances per class.}
\begin{tabular}{|c|c|c|}
\hline
Data    & Genome  & No. of samples \\ \hline
Class-0 & SARS-CoV-2 &4498         \\ \hline
Class-1  & SARS-CoV-1 & 101           \\ \hline
\end{tabular}
\end{table}
\section{Experiments and Results\label{Section_experiments}}
This section deals with the set of experiments evaluated on OCCD and coronavirus genome sequences. We have used  \emph{Python 3}, \emph{LinearSVC}~\cite{fan2008liblinear}, \emph{Numba}~\cite{lam2015numba}, \emph{Numpy}~\cite{harris2020numpy} and {\it Scikit-learn}~\cite{scikit-learn} package for the implementation of ChaosFEX. We compare the performance of ChaosFEX with classical SVM in the low and high training sample regime. Decisions based on limited training samples is a challenging problem for ML algorithms. This learning paradigm with limited samples is referred as few shot learning. Few shot learning aims to develop ML models which generalizes from a small set of labeled training data. There has been previous research in few shot learning~\cite{oreshkin2018tadam},~\cite{qiao2018few}. In this paper, we use ChaosFEX to learn from fewer training samples.
\subsection{Data Preprocessing for Genome Sequence}
Data preprocessing is the first step in any machine learning task. The following data preprocessing is carried out for genome sequence data.

\begin{itemize}
    \item Step 1: Conversion of nucleotide sequence into numeric format. We choose the same numeric conversion as mentioned in~\cite{cnn2020accurate}. The numeric conversion is as follows: Cytosine (C) = 0.25, Thymine (T) = 0.50, Guanine (G) = 0.75, Adenine (A) = 1.0.
    \item Step 2: Compute the absolute value of the Fast Fourier Transform (FFT) of the numeric genome sequence.
    \item Step 3: The FFT coefficients of the nucleotide sequences are normalized independently so as to lie in the range [0,1]. 
\end{itemize}
 \begin{figure*}
\centering
	\begin{subfigure}{0.45\linewidth}
		\centering
		\includegraphics[width=1\textwidth]{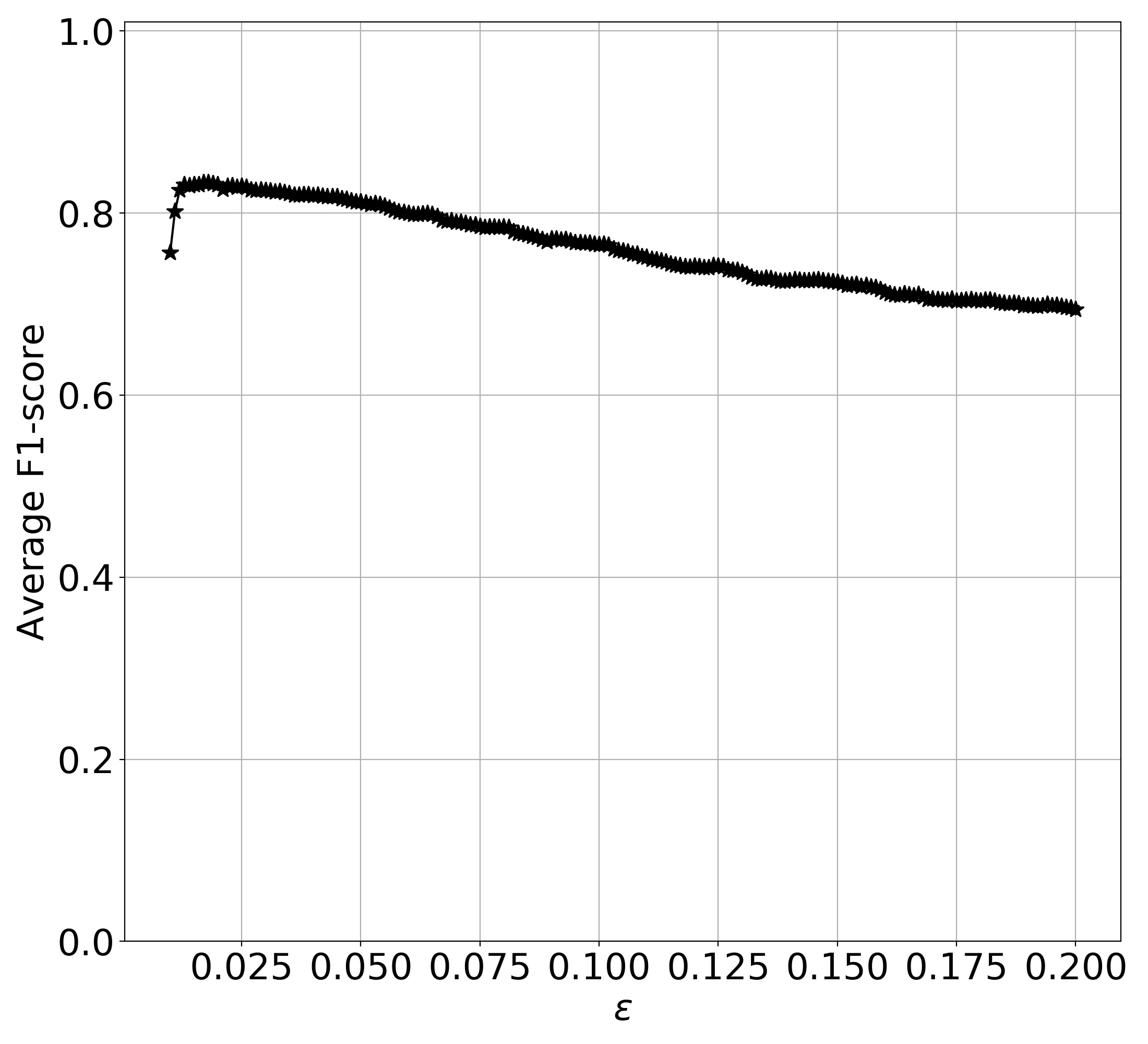}
		\caption{}\label{Fig_occd-hyperparameter-tuning}
	\end{subfigure}
	\begin{subfigure}{0.45\linewidth}
		\centering
		\includegraphics[width=1\linewidth]
{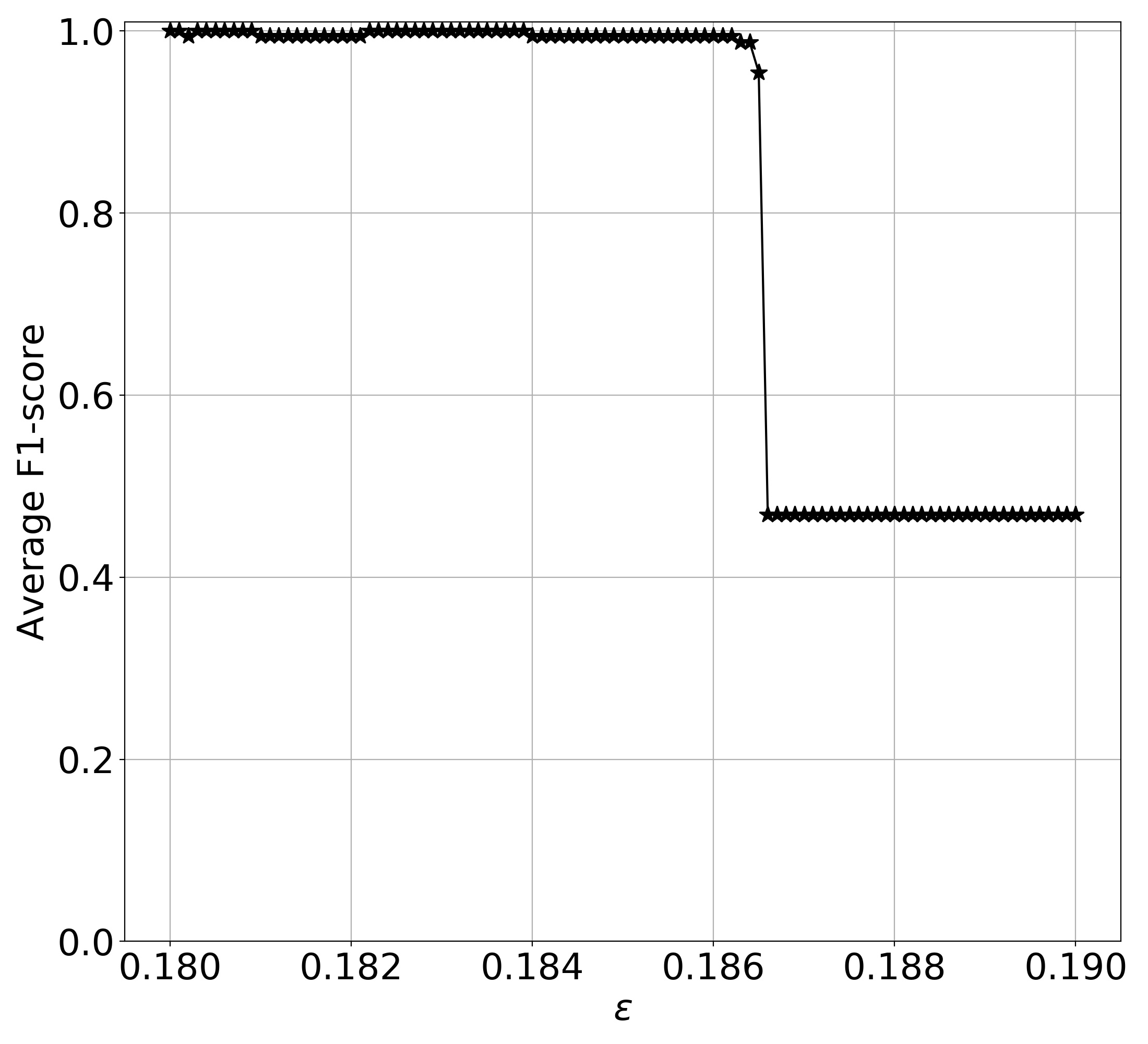}
		\caption{}\label{Fig_binary-class-SARS-CoV-2-vs-others-hyperparameter-tuning}
	\end{subfigure}
    \caption{(\subref{Fig_occd-hyperparameter-tuning}) Hyperparameter tuning for OCCD-Macro averaged F1-score vs. $\epsilon$ for binary classification task (OCCD). $\epsilon$ is varied from $0.01$ to $0.2$ with a step-size of $0.001$. A maximum F1-score of 0.833 was obtained for an $\epsilon = 0.018$ in three fold validation. (\subref{Fig_binary-class-SARS-CoV-2-vs-others-hyperparameter-tuning}) Hyperparameter tuning for SARS-CoV-2 vs. others (Table~\ref{Table_binary_class})-Macro averaged F1-score vs. $\epsilon$ for binary classification task. $\epsilon$ is varied from $0.18$ to $0.19$ with a step-size of $0.0001$. A maximum F1-score of 1.0 was obtained for several values of $\epsilon$ in three fold validation. Out of these several values of $\epsilon$ we choose $\epsilon = 0.183$ as the best hyperparameter.}
\label{fig:3:2}
\end{figure*}%
\subsection{Hyperparameter Tuning}
In this section we discuss the  hyperparameters that were arrived at for NL (ChaosFEX$+$SVM) and SVM. 
\subsubsection{Hyperparameter tuning for NL (ChaosFEX$+$SVM) for OCCD}
The NL (ChaosFEX$+$SVM) has three hyperparameters: initial neural activity ($q$), discrimination threshold ($b$) and epsilon ($\epsilon$). In order to find the best hyperparameters we do three fold crossvalidation. In the case of OCCD, the hyperparameter tuning is as follows: we fixed the initial neural activity $q$ = 0.22, discrimination threshold $b$ = 0.96, and varied $\epsilon$ from $0.01$ to $0.2$ with a step size of $0.001$. In the three fold crossvalidation we get an average macro F1-score of $0.833$ for an $\epsilon = 0.018$. Figure~\ref{Fig_occd-hyperparameter-tuning} represents the results corresponding to hyperparameter tuning.

\subsubsection{Hyperparameter tuning for SVM for OCCD}
We did five fold crossvalidation for the following hyperparameters of SVM-- $kernel =$ [`linear', `rbf'], $gamma =$ [`scale', `auto']. The train and validation split (train(\%), Val (\%)) for fold 1, fold 2, fold 3, fold 4 and fold 5 were chosen to be approximately ($80\%$, $20\%$). We used \emph{LinearSVC}~\cite{fan2008liblinear} for the implementation of SVM with linear kernel and \emph{LIBSVM}~\cite{chang2011libsvm} for the implementation of SVM with RBF kernel. We got a maximum average macro F1-score of 0.837 using  SVM with RBF kernel for  $gamma =$`scale'.

\subsubsection{Hyperparameter tuning for NL (ChaosFEX$+$SVM) for binary classification (SARS-CoV-2 vs. others) dataset}
We fixed the initial neural activity $q$ as $0.34$, discrimination threshold $b$ as $0.499$. For this $q$ and $b$, we varied the epsilon $\epsilon$ from $0.18$ to $0.19$ with a step size of $0.0001$. We did a three fold crossvalidation. In each of the three folds we took 66.6\% data for training and 33\% for validating. We got an average F1-score of 1, for $\epsilon$ in the range starting from $0.18$ to $0.1839$ with a step size of $0.0001$ in the three fold validation. Out of these values, we choose $\epsilon = 0.183$ for further analysis. Figure~\ref{Fig_binary-class-SARS-CoV-2-vs-others-hyperparameter-tuning} shows the graph of macro averaged F1-score of threefold crossvalidation vs. $\epsilon$. 
\subsubsection{Hyperparameter tuning for SVM for binary classification (SARS-CoV-2 vs. others) dataset} 
We did a five fold crossvalidation for the following hyperparameters of SVM: $kernel =$ [`linear', `rbf'], $gamma =$ [`scale', `auto']. The train and validation split (train(\%), Val (\%)) for fold 1, fold 2, fold 3, fold 4 and fold 5 were chosen to be approximately ($80\%$, $20\%$). We got a maximum average macro F1-score of 1 using linear kernel. SVM with RBF kernel for  $gamma =$`scale' also gave a maximum average macro F1-score of 1. We use $kernel$ = `linear' for further experiments.

\begin{figure}[!h]
\centering
	\begin{subfigure}{0.35\linewidth}
		\centering
		\includegraphics[width=1\linewidth]{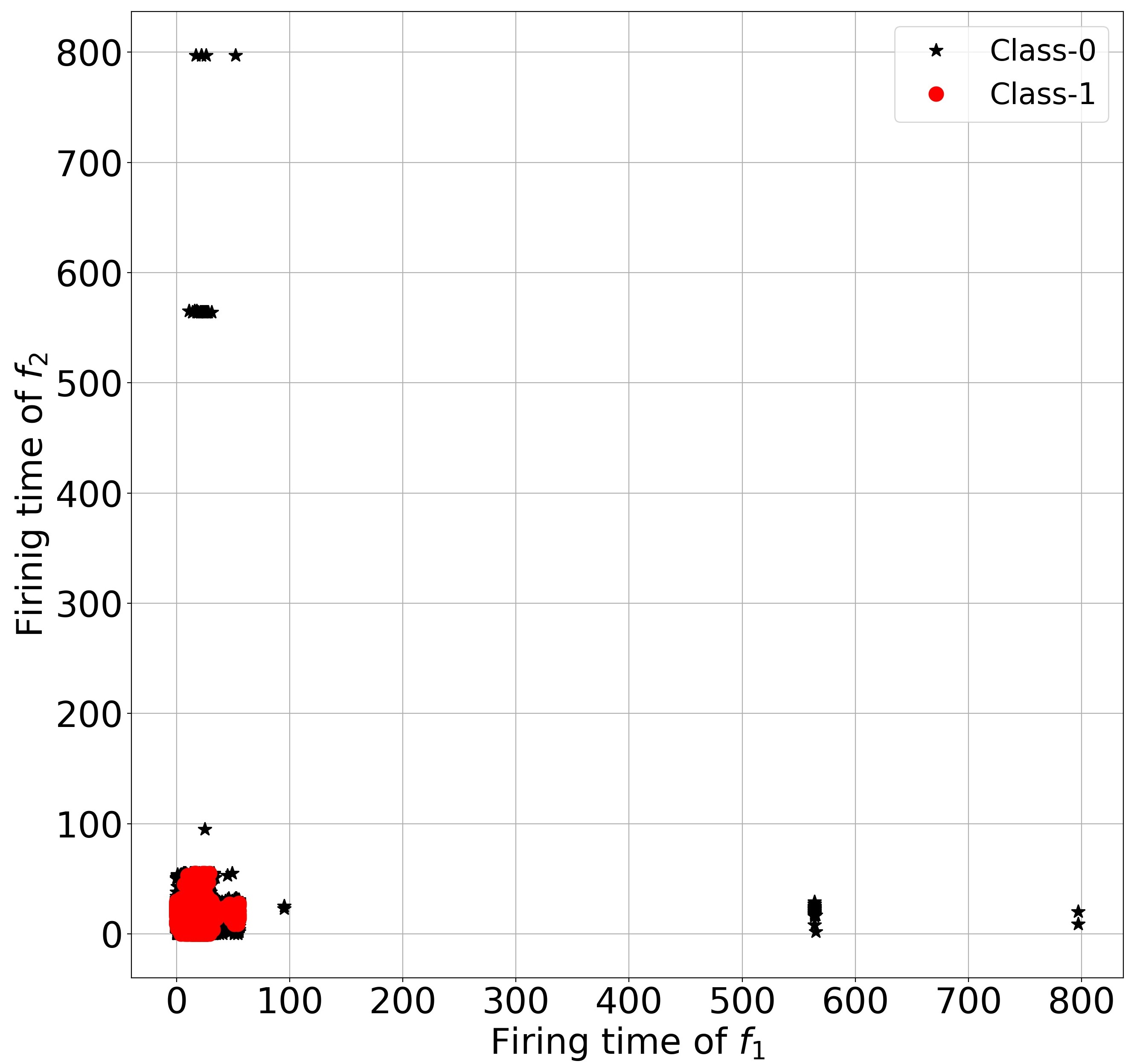}
		\caption{}\label{Fig_occd_train_data_firing_time}
	\end{subfigure}
	\begin{subfigure}{0.35\linewidth}
		\centering
		\includegraphics[width=1\linewidth]{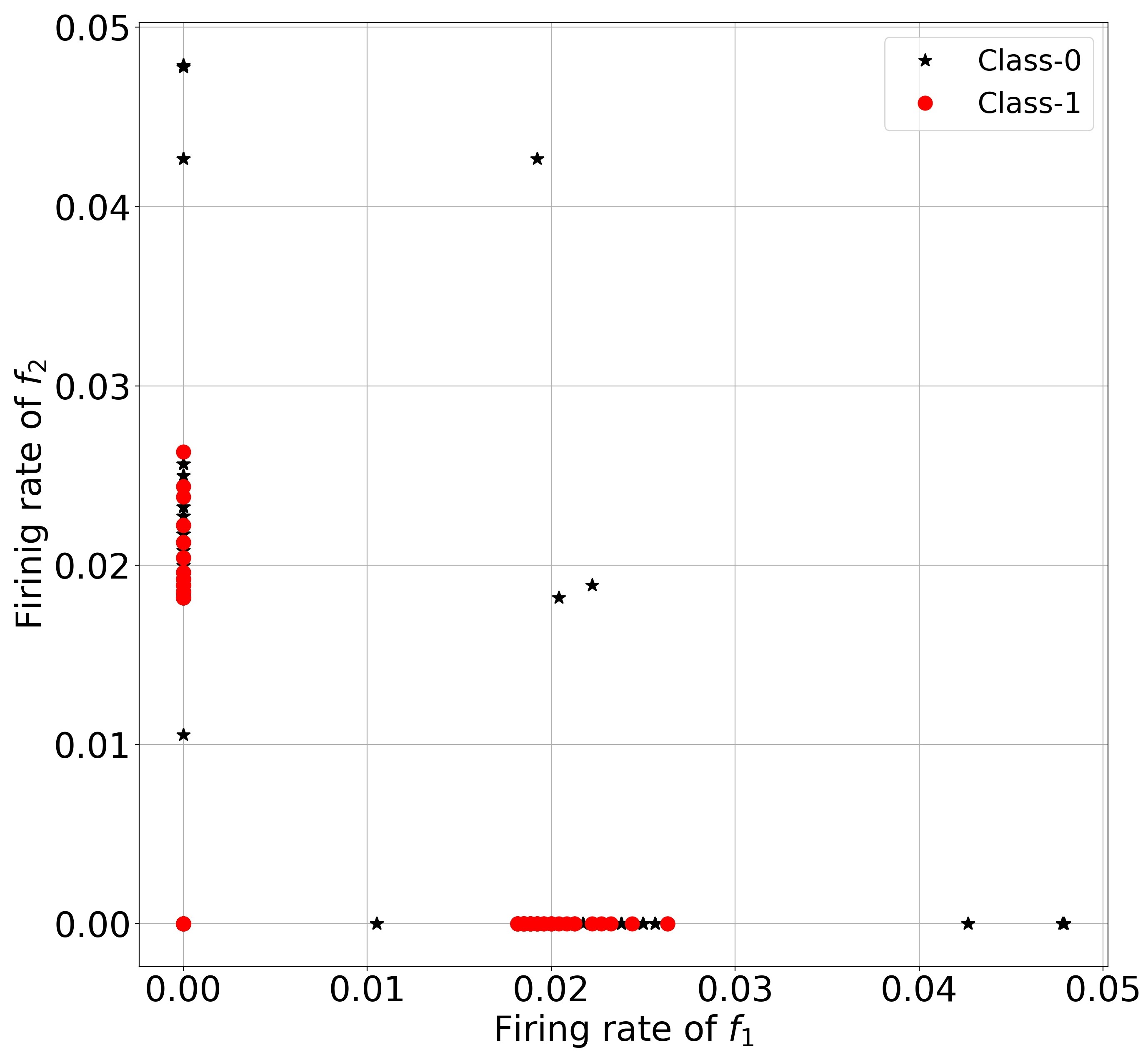}
		\caption{}\label{Fig_occd_train_data_firing_rate}
	\end{subfigure}
	\begin{subfigure}{0.35\linewidth}
		\centering
		\includegraphics[width=1\linewidth]{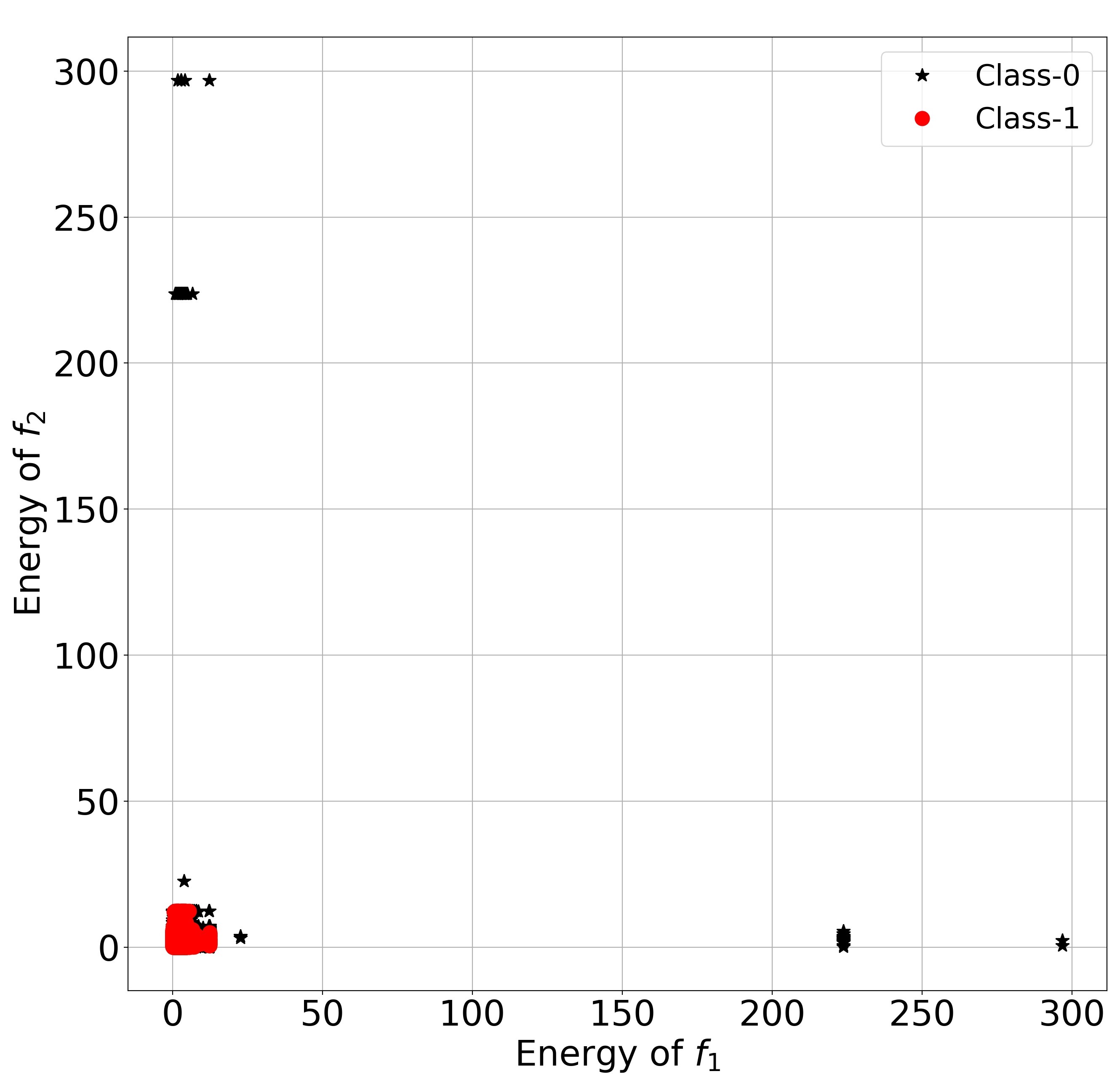}
		\caption{}\label{Fig_occd_train_data_energy}
	\end{subfigure}
		\begin{subfigure}{0.35\linewidth}
		\centering
		\includegraphics[width=1\linewidth]{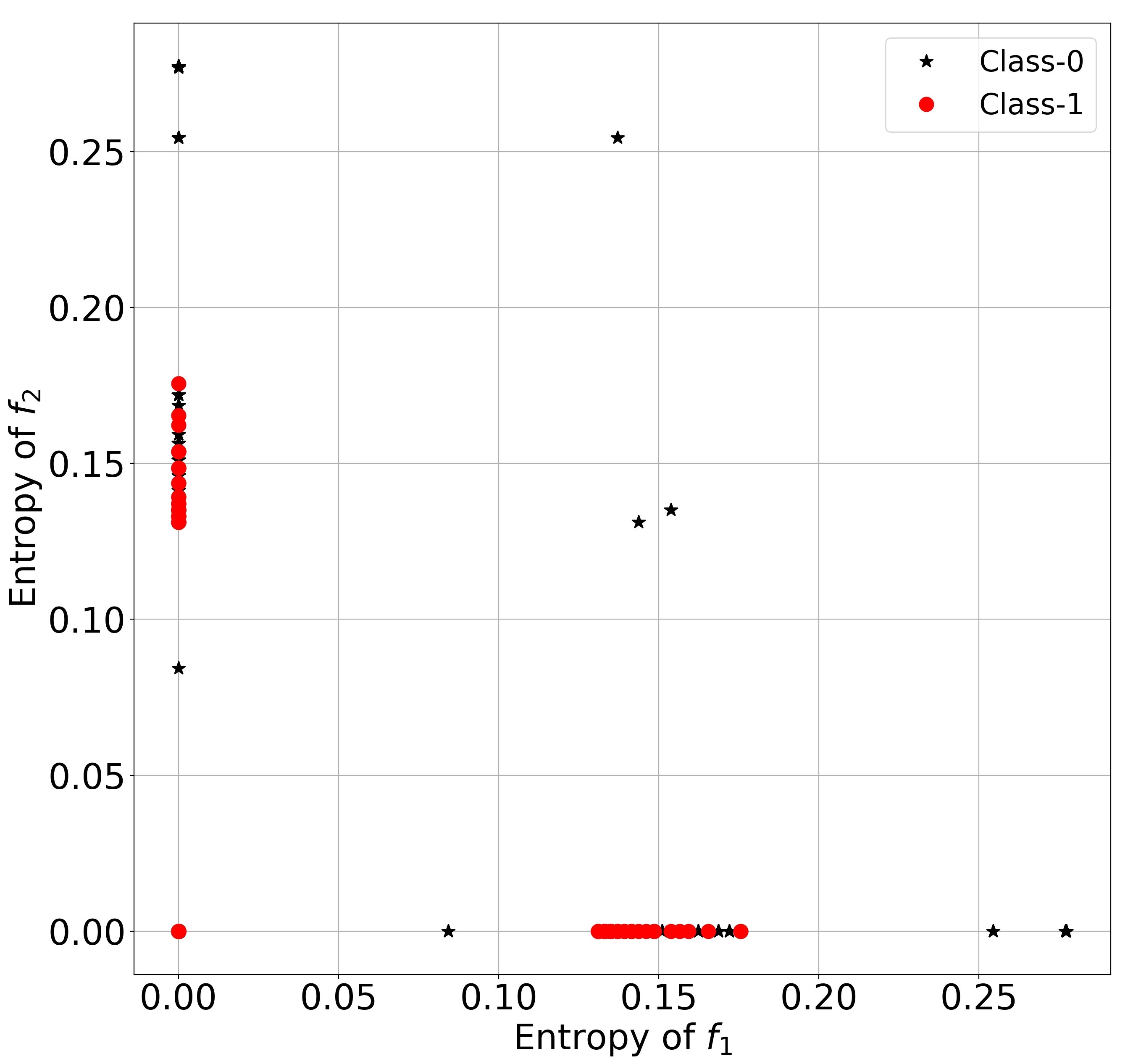}
		\caption{}\label{Fig_occd_train_data_entropy}
	\end{subfigure}
	\caption{ChaosFEX features for OCCD. (\subref{Fig_occd_train_data_firing_time}) Firing time corresponding to the normalized train data. (\subref{Fig_occd_train_data_firing_rate}) Firing rate corresponding to the normalized train data. (\subref{Fig_occd_train_data_energy}) Energy of the chaotic trajectory corresponding to the normalized train data. (\subref{Fig_occd_train_data_entropy}) Entropy of the chaotic trajectory corresponding to the normalized train data. }
\end{figure}%
\subsection{OCCD Results}
We performed two sets of experiments on the OCCD. The first experiment uses the train distribution provided in Table~\ref{Table_OCCD_train_test_split}, and evaluated the performance of ChaosFEX and SVM with RBF kernel on the test data. The second experiment deals with low training sample regime. In both these experiments we compare the performance of ChaosFEX  with SVM (RBF kernel). The input data passed to ChaosFEX and SVM with RBF kernel are the same.

\subsubsection{OCCD: Performance on testdata}
The accuracy, macro averaged precision, recall and F1-score are provided in Table~\ref{Table_OCCD_results}. The classification metric definitions are provided in~\cite{classification_metric}. 

In this task, NL (ChaosFEX+SVM) slightly improved the performance when compared to the input data directly passed to SVM with RBF kernel. In the chaotic feature space, SVM with linear kernel was able to give a slightly improved performance. This empirically shows the effectiveness of chaotic feature space. The chaotic feature space (Firing time, Firing rate, Energy of the chaotic trajectory, Entropy of the symbolic sequence of the chaotic trajectory) can be seen as an efficient feature engineering technique.

\begin{table}[!h]
\centering
\caption{Prediction for OCCD. `Neurochaos' inspired hybrid architecture (NL) with 4 ChaosFEX features (Firing rate, Energy, Firing time, Entropy) followed by SVM classifier with linear kernel is compared with SVM with RBF kernel.}
\begin{tabular}{|c|c|c|}
\hline
Data & \multicolumn{2}{c|}{OCCD} \\ \hline
Method & \begin{tabular}[c]{@{}c@{}}SVM\\ (RBF)\end{tabular} & ChaosFEX+SVM \\ \hline
Accuracy & 83.10 & 83.61 \\ \hline
Precision & 0.83 & 0.84 \\ \hline
Recall & 0.83 & 0.84 \\ \hline
F1-score & 0.83 & 0.84 \\ \hline
 \end{tabular}
\label{Table_OCCD_results}
\end{table}

\subsubsection{Chaotic Feature Space for OCCD traindata}
In the extracted feature space (Figure~\ref{Fig_occd_train_data_firing_time}, Figure~\ref{Fig_occd_train_data_firing_rate}, Figure~\ref{Fig_occd_train_data_energy} and Figure~\ref{Fig_occd_train_data_entropy}) of train data, the data belonging to class-0 and class-1 are populated in different clusters. The data is not spread everywhere in the extracted feature space. There are multiple clusters of data belonging to class-0 and class-1 in the extracted feature space. In Figure~\ref{Fig_occd_train_data_firing_time} and Figure~\ref{Fig_occd_train_data_energy}, the class-1 firing time features and energy features are grouped in one side. Whereas the class-0 firing time and energy features are grouped as multiple clusters in different direction.

There is a direct correlation between firing time (Figure~\ref{Fig_occd_train_data_firing_time}) and energy of the chaotic trajectory (Figure~\ref{Fig_occd_train_data_energy}). As firing time increases, the length of the chaotic trajectory also increases. As length increases the energy of the chaotic trajectory also increases. This is because energy is defined as the square of the $l_{2}$-norm of chaotic trajectory. Hence firing time and energy of the chaotic trajectory are positively correlated.  Similarly, the feature extracted space of Firing rate (Figure~\ref{Fig_occd_train_data_firing_rate}) and Entropy of the symbolic sequence of the chaotic trajectory (Figure~\ref{Fig_occd_train_data_entropy}) have similar patterns. These four features together is yielding a macro averaged F1-score of 0.84 on test data.
\begin{figure*}
\centering
	\begin{subfigure}{0.4\linewidth}
		\centering
		\includegraphics[width=1\linewidth]{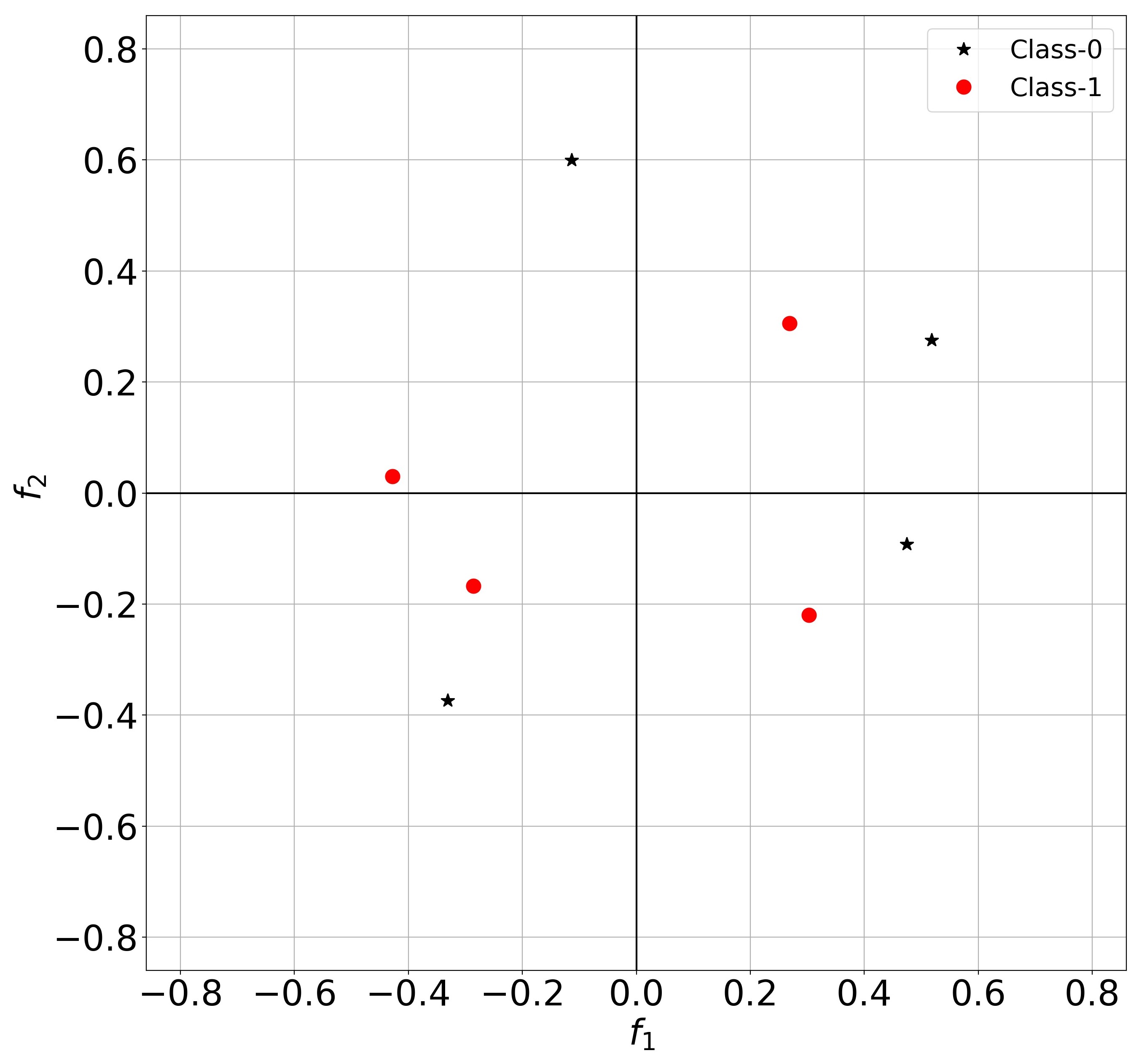}
		\caption{}\label{Fig_occd_train_data_4_samples}
	\end{subfigure}
	\begin{subfigure}{0.4\linewidth}
		\centering
		\includegraphics[width=1\linewidth]{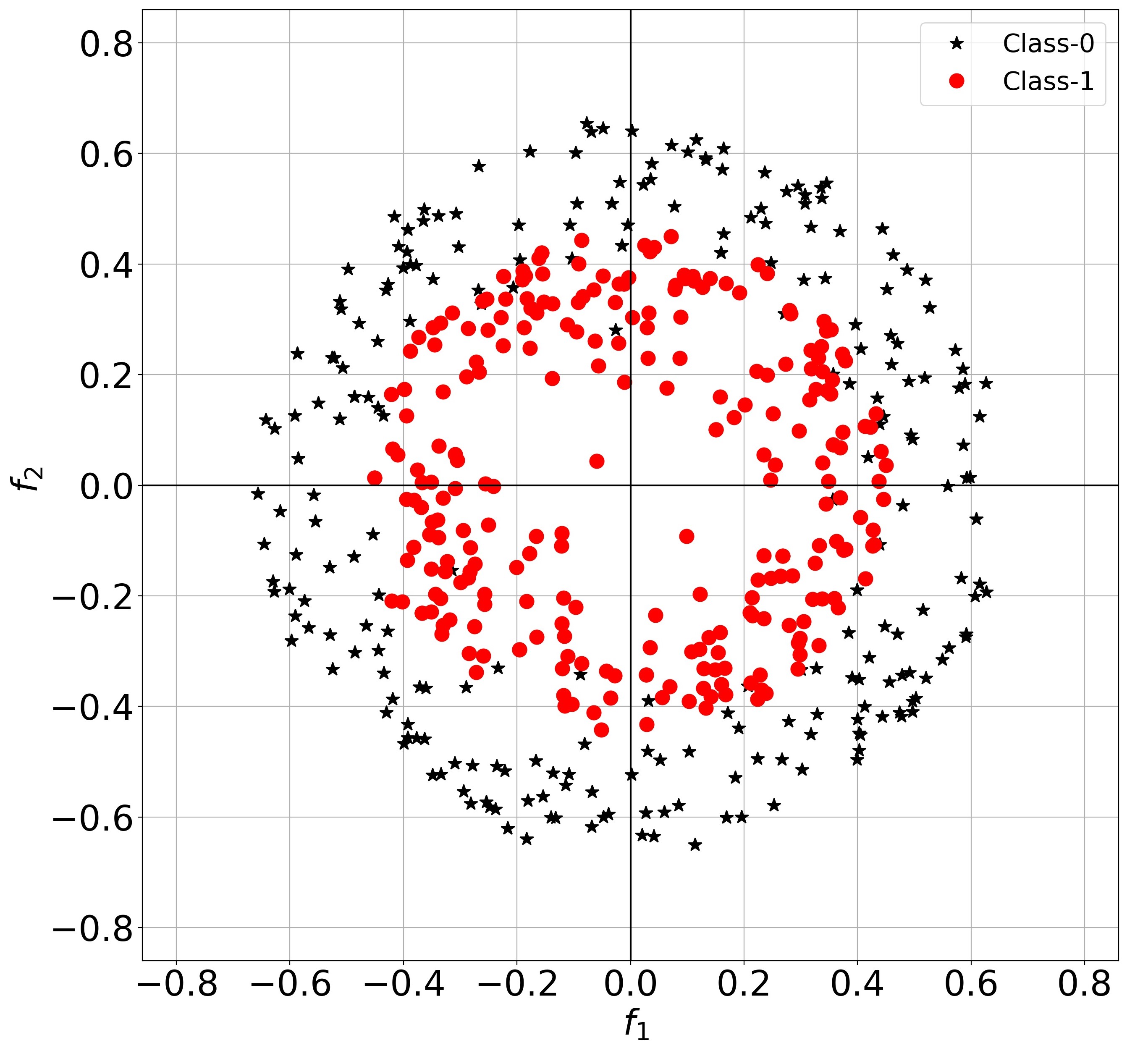}
		\caption{}\label{Fig_occd_train_data_260_samples}
	\end{subfigure}
	\begin{subfigure}{0.4\linewidth}
		\centering
		\includegraphics[width=1\linewidth]{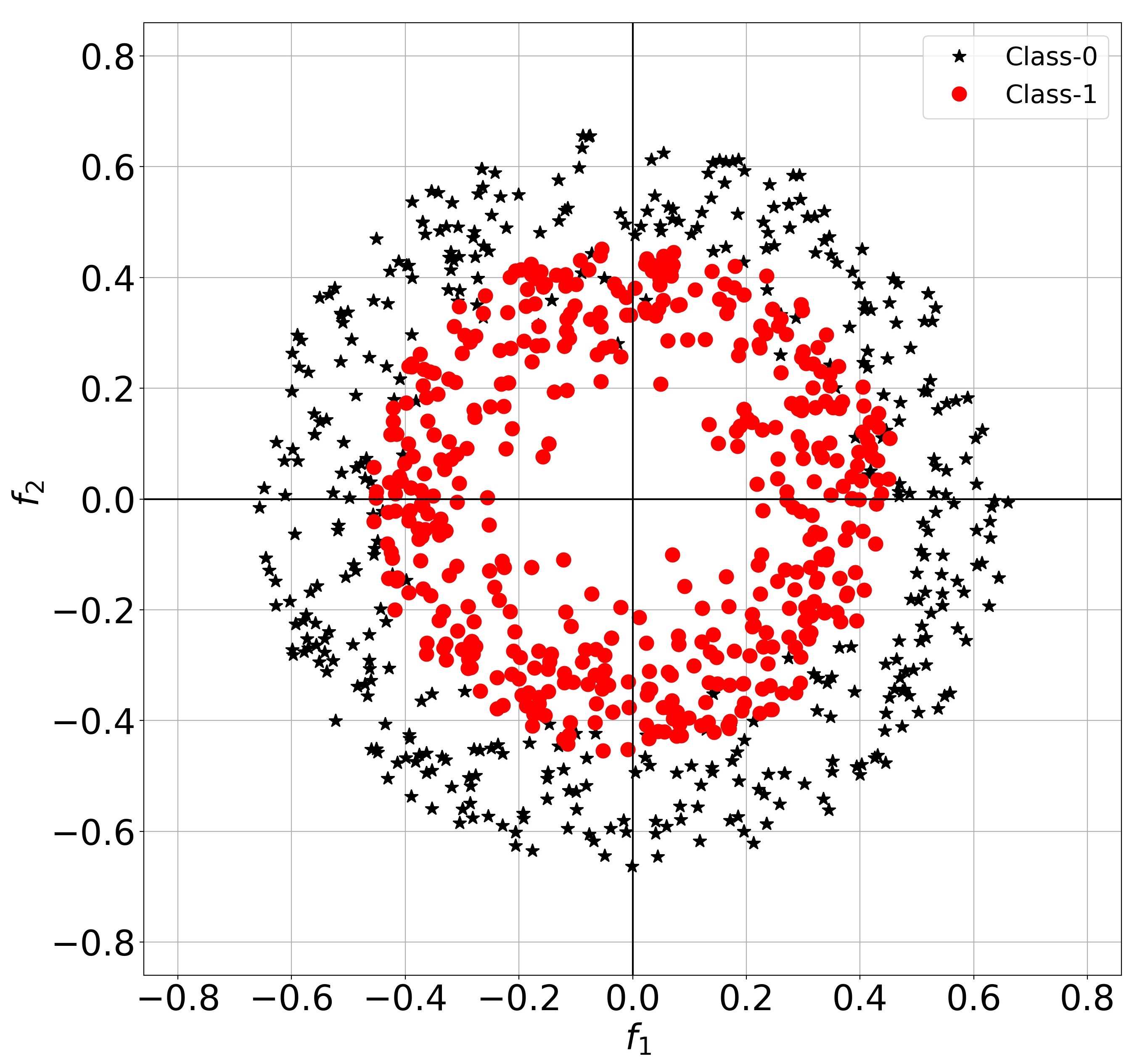}
		\caption{}\label{Fig_occd_train_data_500_samples}
	\end{subfigure}
		\begin{subfigure}{0.4\linewidth}
		\centering
		\includegraphics[width=1\linewidth]{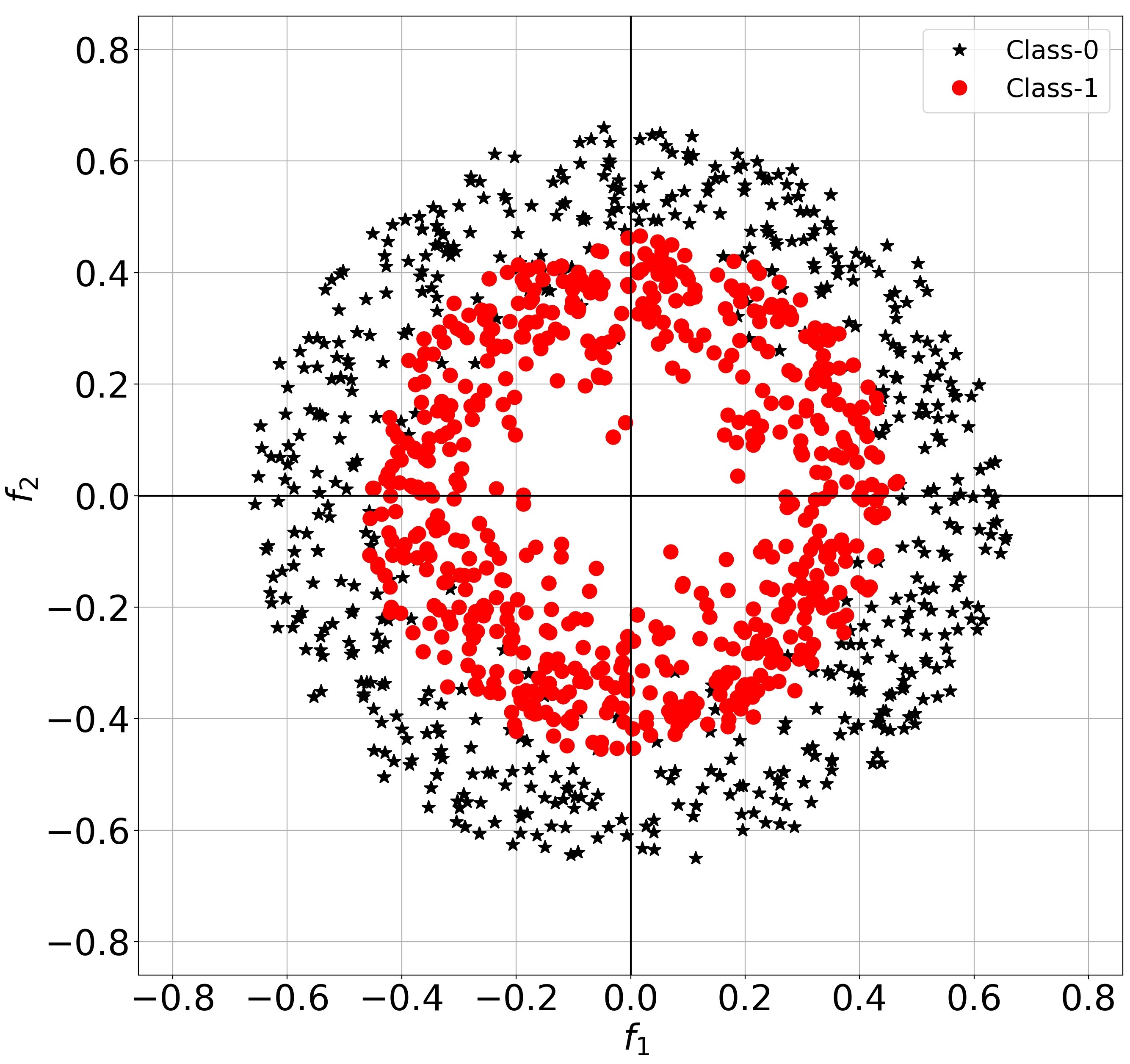}
		\caption{}\label{Fig_occd_train_data_724_samples}
	\end{subfigure}
\caption{OCCD training samples. (\subref{Fig_occd_train_data_4_samples})  Train data with 4 samples per class. (\subref{Fig_occd_train_data_260_samples}) Train data with 260 samples per class. (\subref{Fig_occd_train_data_500_samples}) Train data with 500 samples per class. (\subref{Fig_occd_train_data_724_samples}) Train data with 724 samples per class.}
\label{fig:3:2}
\end{figure*}%
\begin{figure*}
    \centerline{ \includegraphics[width=1\textwidth]{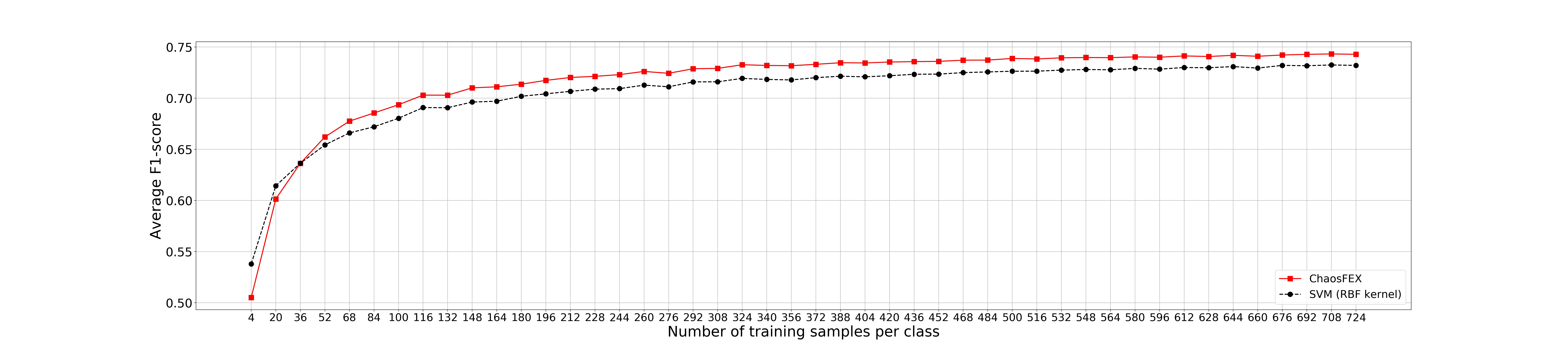}}
    \caption{OCCD Low training sample regime: Average macro F1-score for 200 random trials of training with $4, 20, 36,\ldots, 724$ training samples per class. In the low training sample regime ChaosFEX slightly outperforms SVM with RBF kernel from training with 36 sample per class onwards. ChaosFEX implies ChaosFEX+SVM (linear) in all plots.}
    \label{Fig_occd-lts-f1score_vs_num_samples}
    \end{figure*}
 \begin{figure*}
    \centerline{ \includegraphics[width=1\textwidth]{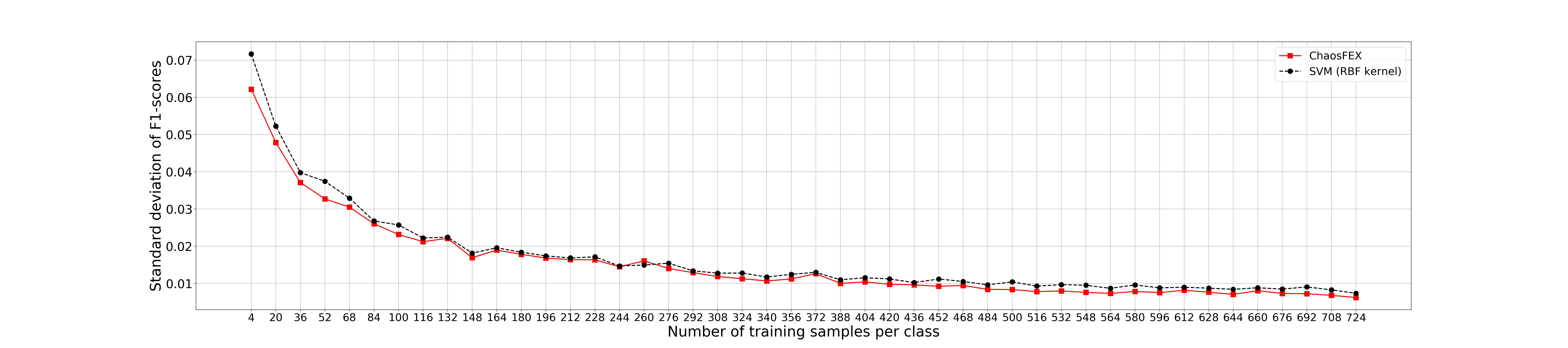}}
    \caption{OCCD Low training sample regime: Standard deviation of  macro F1-scores for 200 random trials of training with $4, 20, 36,\ldots, 724$ training samples per class. The standard deviation of ChaosFEX is slightly lower compared to SVM with RBF kernel. }
    \label{Fig_occd-lts-std-f1score_vs_num_samples}
    \end{figure*}
\subsubsection{OCCD: Low training sample regime}
One shot learning or few shot learning is one of the challenging problems in machine learning. In many tasks a huge amount of training data will not be available. For example, during the time of an initial outbreak of a pandemic, decisions has to be made from fewer number of samples. In such a situation learning from limited samples plays a key role in controlling the spread of pandemic disease. We evaluate the efficacy of the proposed method in the low training sample regime. The hyperparameters used are the same for low training sample regime as well as high training sample regime ($q = 0.22, b = 0.96, \epsilon = 0.018$). 
In the low training sample regime we did not consider 30\% of data belonging to each of the four quadrants for both class-0 and class-1. The  30\% removal of data instances is based on the length of the data. The length of each data point ($f_i,f_j$) is $\sqrt{f_i^{2} + f_j^{2}}$. The topmost 30\% of the data instances (in each quadrant) with highest lengths were removed. We did this in order to avoid too many overlapping data instances between class-0 and class-1 data instances. After removal of the $30 \%$ of the data instances from each quadrant, we get a reduced dataset which is used subsequently for training. The test dataset is unchanged (refer to Table~\ref{Table_OCCD_train_test_split}).

We did 200 random trials of training with $4, 20, 36, \ldots, 724$ (step size = $16$) samples per class. These samples are drawn uniformly from the four quadrants. Figure~\ref{Fig_occd_train_data_4_samples}, \ref{Fig_occd_train_data_260_samples}, \ref{Fig_occd_train_data_500_samples} and \ref{Fig_occd_train_data_724_samples} depict the training data for four cases in the low training sample regime. We then found macro F1-score for the test set provided in Table~\ref{Table_OCCD_train_test_split}. Since we did 200 random trails of training, we have 200 macro F1-scores for training with $4, 20, \ldots, 724$ samples per class. We computed the mean F1-score for the 200 random trials of training with $4, 20, \dots 724$. Figure~\ref{Fig_occd-lts-f1score_vs_num_samples} and Figure~\ref{Fig_occd-lts-std-f1score_vs_num_samples} represents average macro F1-score vs. the number of training samples per class and standard deviation of F1-scores vs. number of training samples per class respectively. ChaosFEX consistently performs slightly better than SVM (RBF kernel) when the number of training samples per class is 36 or higher.
 %
\subsection{Binary Classification (SARS-CoV-2 vs. others)}
The data corresponding to the classification of SARS-CoV-2 from other viruses are provided in Table~\ref{Table_binary_class}. Because the data was highly imbalanced, we carried out a five fold stratified crossvalidation on the binary classification (SARS-CoV-2 vs. Others) data. The train and validation split (train (\%), Val (\%)) for  fold 1, fold 2, fold 3, fold 4 and fold 5 were chosen to be approximately ($80\%$, $20\%$). The maximum length of the genome sequence is 31029. For each fold we calculated precision, recall and F1-score separately. Using this information we computed the macro averaged precision (Pr), recall (Re) and F1-score (F1) for that particular fold~\cite{classification_metric}. The results using ChaosFEX for the five fold crossvalidation is provided in Table~\ref{Table_binary_classification_five_fold_validation_neurochaos}. Table~\ref{Table_binary_classification_five_fold_validation_svm} represents the five fold crossvalidation for the same data distribution for SVM with linear kernel (without ChaosFEX features). For both NL (ChaosFEX+SVM) and SVM with linear kernel, we get an average macro F1-score of 1.0 in five fold stratified crossvalidation.
\begin{table*}
\centering
\caption{\label{Table_binary_classification_five_fold_validation_neurochaos} SARS-CoV-2 vs. others: Five fold crossvalidation for binary classification problem using ChaosFEX+SVM.}
\begin{tabular}{|c|c|c|c|c|c|c|c|c|c|c|c|c|c|c|c|}
\hline
Folds & \multicolumn{3}{c|}{Fold - 1} & \multicolumn{3}{c|}{Fold - 2} & \multicolumn{3}{c|}{Fold - 3} & \multicolumn{3}{c|}{Fold - 4} & \multicolumn{3}{c|}{Fold - 5} \\ \hline
Metric & Pr & Re & F1 & Pr & Re & F1 & Pr & Re & F1 & Pr & Re & F1 & Pr & Re & F1 \\ \hline
Class-0 & 1 & 1 & 1 & 1 & 1 & 1 & 1 & 1 & 1 & 1 & 1 & 1 & 1 & 1 & 1 \\ \hline
Class-1 & 1 & 1 & 1 & 1 & 1 & 1 & 1 & 1 & 1 & 1 & 1 & 1 & 1 & 1 & 1 \\ \hline
\begin{tabular}[c]{@{}c@{}}Macro \\ Averaged\end{tabular} & 1 & 1 & 1 & 1 & 1 & 1 & 1 & 1 & 1 & 1 & 1 & 1 & 1 & 1 & 1 \\ \hline
\end{tabular}
\end{table*}
\begin{table}[!h]
\centering
\caption{\label{Table_binary_classification_five_fold_validation_svm} SARS-CoV-2 vs. others: Five fold crossvalidation for binary classification problem using SVM with linear kernel.}
\begin{tabular}{|c|c|c|c|c|c|c|c|c|c|c|c|c|c|c|c|}
\hline
Folds & \multicolumn{3}{c|}{Fold - 1} & \multicolumn{3}{c|}{Fold - 2} & \multicolumn{3}{c|}{Fold - 3} & \multicolumn{3}{c|}{Fold - 4} & \multicolumn{3}{c|}{Fold - 5} \\ \hline
Metric & Pr & Re & F1 & Pr & Re & F1 & Pr & Re & F1 & Pr & Re & F1 & Pr & Re & F1 \\ \hline
Class-0 & 1 & 1 & 1 & 1 & 1 & 1 & 1 & 1 & 1 & 1 & 1 & 1 & 1 & 1 & 1 \\ \hline
Class-1 & 1 & 1 & 1 & 1 & 1 & 1 & 1 & 1 & 1 & 1 & 1 & 1 & 1 & 1 & 1 \\ \hline
\begin{tabular}[c]{@{}c@{}}Macro \\ Averaged\end{tabular} & 1 & 1 & 1 & 1 & 1 & 1 & 1 & 1 & 1 & 1 & 1 & 1 & 1 & 1 & 1 \\ \hline
\end{tabular}
\end{table}
\begin{table*}[!h]
\centering
\caption{\label{Table_multi_class_classification_five_fold_validation_neurochaos} Multiclass Classification: Five fold crossvalidation for multiclass classification problem using ChaosFEX.}
\begin{tabular}{|c|c|c|c|c|c|c|c|c|c|c|c|c|c|c|c|}
\hline
Folds & \multicolumn{3}{c|}{Fold - 1} & \multicolumn{3}{c|}{Fold - 2} & \multicolumn{3}{c|}{Fold - 3} & \multicolumn{3}{c|}{Fold - 4} & \multicolumn{3}{c|}{Fold - 5} \\ \hline
Metric & Pr & Re & F1 & Pr & Re & F1 & Pr & Re & F1 & Pr & Re & F1 & Pr & Re & F1 \\ \hline
Class-0 & 1 & 1 & 1 & 1 & 1 & 1 & 1 & 0.92 & 0.96 & 1 & 1 & 1 & 1 & 1 & 1 \\ \hline
Class-1 & 1 & 1 & 1 & 1 & 1 & 1 & 1 & 1 & 1 & 1 & 1 & 1 & 1 & 1 & 1 \\ \hline
Class-2 & 1 & 1 & 1 & 1 & 1 & 1 & 1 & 1 & 1 & 1 & 1 & 1 & 1 & 1 & 1 \\ \hline
Class-3 & 1 & 1 & 1 & 1 & 1 & 1 & 0.94 & 1 & 0.97 & 1 & 1 & 1 & 1 & 1 & 1 \\ \hline
Class-4 & 1 & 1 & 1 & 1 & 1 & 1 & 1 & 1 & 1 & 1 & 1 & 1 & 1 & 1 & 1 \\ \hline
\begin{tabular}[c]{@{}c@{}}Macro \\ Averaged\end{tabular} & 1 & 1 & 1 & 1 & 1 & 1 & 0.99 & 0.98 & 0.99 & 1 & 1 & 1 & 1 & 1 & 1 \\ \hline
\end{tabular}
\end{table*}
\begin{table*}[!h]
\centering
\caption{\label{Table_multi_class_classification_five_fold_validation_svm} Multiclass Classification: Five fold crossvalidation for multiclass classification problem using SVM with linear kernel.}
\begin{tabular}{|c|c|c|c|c|c|c|c|c|c|c|c|c|c|c|c|}
\hline
Folds & \multicolumn{3}{c|}{Fold - 1} & \multicolumn{3}{c|}{Fold - 2} & \multicolumn{3}{c|}{Fold - 3} & \multicolumn{3}{c|}{Fold - 4} & \multicolumn{3}{c|}{Fold - 5} \\ \hline
Metric & Pr & Re & F1 & Pr & Re & F1 & Pr & Re & F1 & Pr & Re & F1 & Pr & Re & F1 \\ \hline
Class-0 & 1 & 1 & 1 & 1 & 1 & 1 & 0.86 & 0.92 & 0.89 & 1 & 1 & 1 & 1 & 1 & 1 \\ \hline
Class-1 & 1 & 1 & 1 & 1 & 1 & 1 & 1 & 1 & 1 & 1 & 1 & 1 & 1 & 1 & 1 \\ \hline
Class-2 & 1 & 1 & 1 & 1 & 1 & 1 & 1 & 1 & 1 & 1 & 1 & 1 & 1 & 1 & 1 \\ \hline
Class-3 & 1 & 1 & 1 & 1 & 1 & 1 & 0.94 & 1 & 0.97 & 1 & 1 & 1 & 1 & 1 & 1 \\ \hline
Class-4 & 1 & 1 & 1 & 1 & 1 & 1 & 0 & 0 & 0 & 1 & 1 & 1 & 1 & 1 & 1 \\ \hline
\begin{tabular}[c]{@{}c@{}}Macro \\ Averaged\end{tabular} & 1 & 1 & 1 & 1 & 1 & 1 & 0.76 & 0.78 & 0.77 & 1 & 1 & 1 & 1 & 1 & 1 \\ \hline
\end{tabular}
\end{table*}
\subsection{Multiclass Classification}
In the case of multiclass classification (Table~\ref{Table_multi_class}), we did a five fold stratified crossvalidation. The train and validation split (train (\%), Val (\%)) for  fold 1, fold 2, fold 3, fold 4 and fold 5 were chosen to be approximately ($80\%$, $20\%$). The maximum length of the genome sequence is 31029. In all the five folds, ChaosFEX slightly outperformed SVM with linear kernel. It is important to highlight that the hyperparameters used in ChaosFEX are the same for multiclass as well as binary class.
With the same hyperparameters, ChaosFEX gave high performance. The hyperparameters are robust to the addition of more classes to this particular existing data. When new classes are added, there is a tendency for ANNs to forget previously learned tasks. This is called catastrophic forgetting~\cite{kirkpatrick2017overcoming}. In our experiments, using the same hyperparameters ChaosFEX preserves the performance of previously learned tasks. This is an empirical evidence of the robustness of ChaosFEX to the catastrophic forgetting problem. The robustness of chaos based machine learning algorithm to catastrophic forgetting problem has been shown in~\cite{laleh2020chaotic}. The results for five fold crossvalidation for ChaosFEX and SVM with linear kernel (without ChaosFEX features) are provided in Table~\ref{Table_multi_class_classification_five_fold_validation_neurochaos} and Table~\ref{Table_multi_class_classification_five_fold_validation_svm}.
%
\begin{figure*}
\centering
	\begin{subfigure}{0.45\linewidth}
		\centering
		\includegraphics[width=1\linewidth]{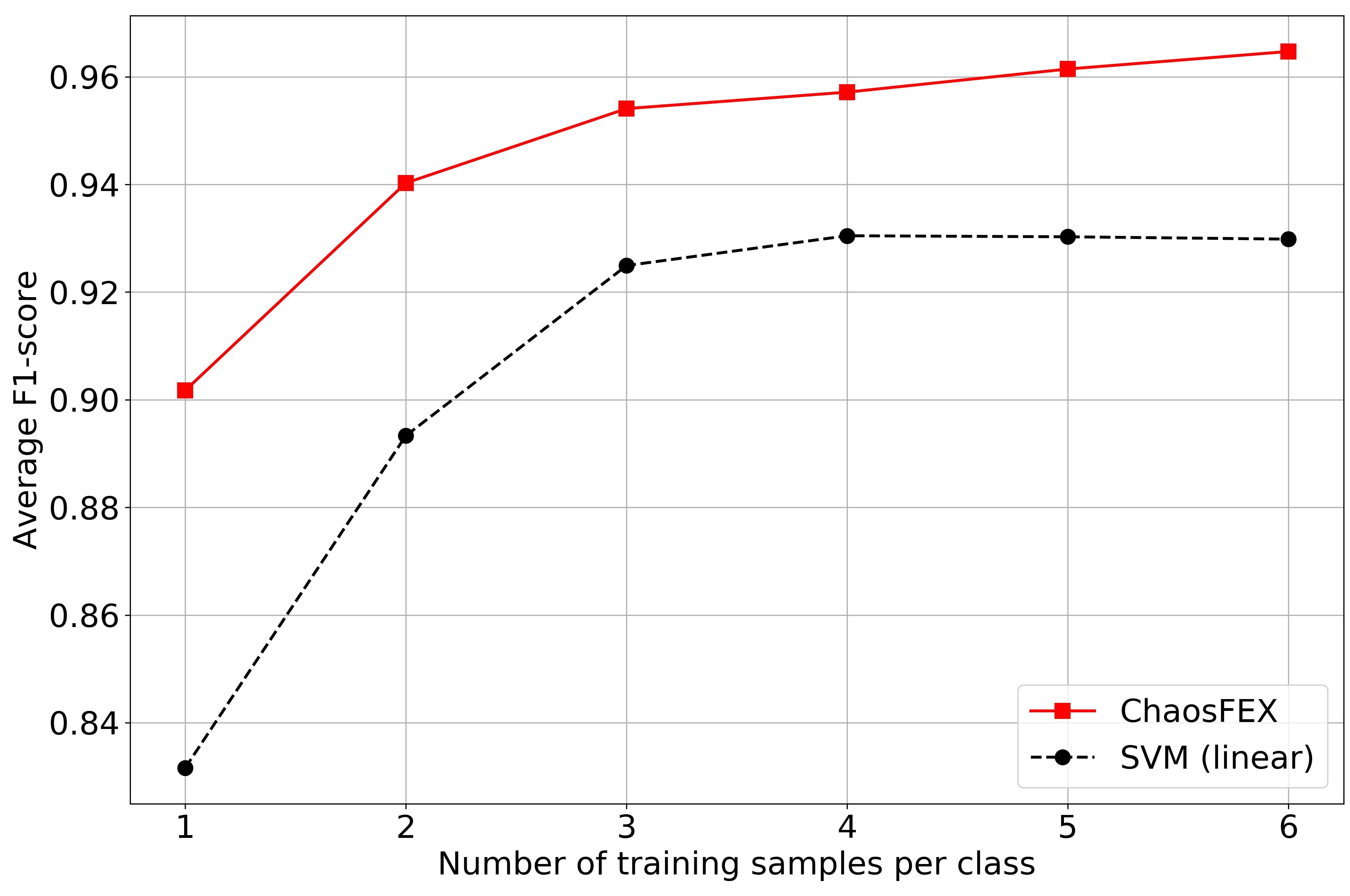}
		\caption{}\label{Fig_F1_low_training_sample_regime_multi_class_classification}
	\end{subfigure}
	\begin{subfigure}{0.45\linewidth}
		\centering
		\includegraphics[width=1\linewidth]{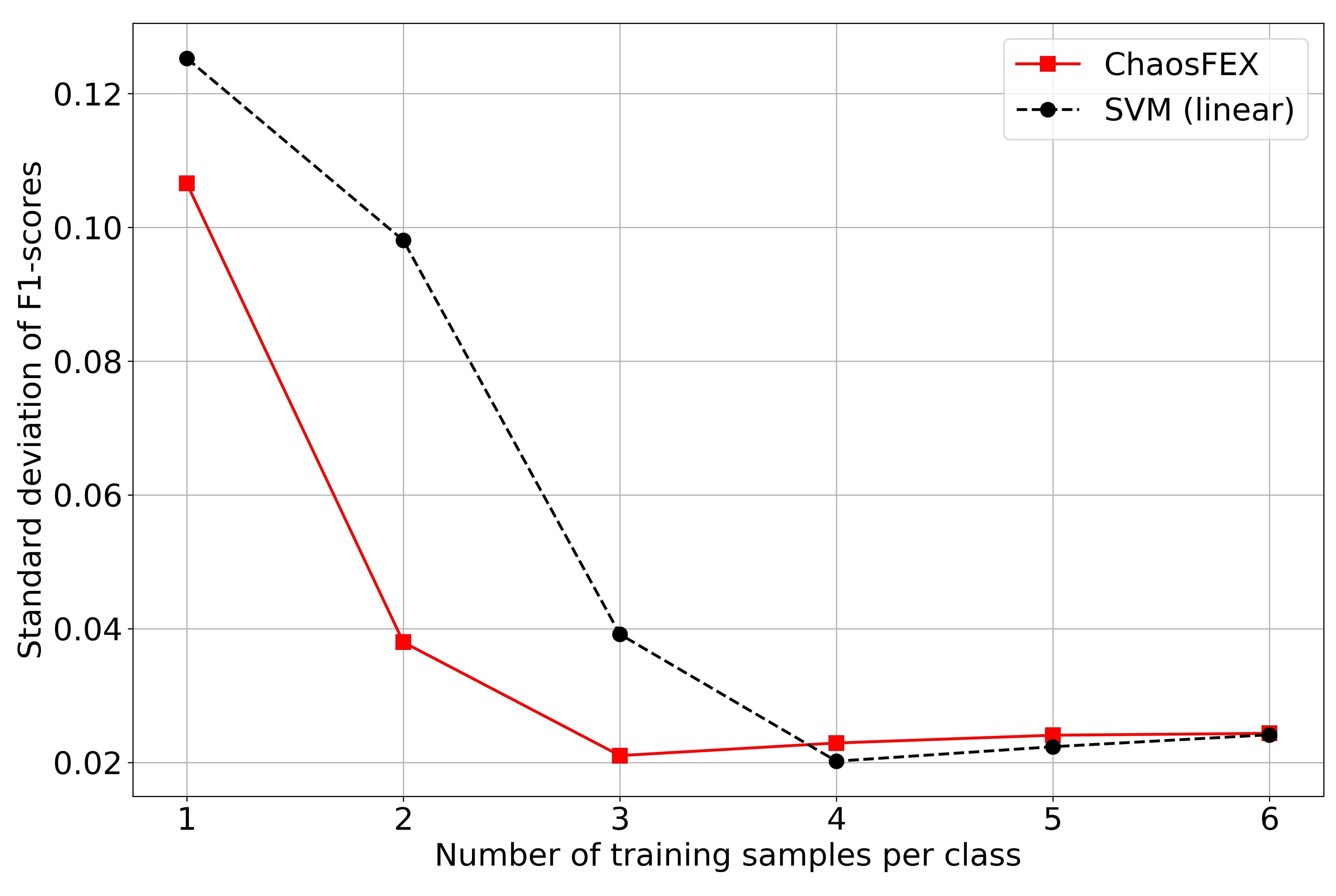}
		\caption{}\label{Fig_standard_deviation_low_training_sample_regime_multi_class_classification}
	\end{subfigure}
\caption{Multiclass low training sample regime. (\subref{Fig_F1_low_training_sample_regime_multi_class_classification}) Average macro F1-score of test data for 200 independent trials of training vs. number of training samples per class for multiclass classification. (\subref{Fig_standard_deviation_low_training_sample_regime_multi_class_classification}) Standard deviation of macro F1-scores for 200 independent trials of training with $1, 2, \ldots, 6$ samples per class for multiclass classification.}
	\label{fig:3:2}
\end{figure*}%
\subsubsection{Multiclass: Low Training Sample Regime}
In the low training sample regime we used $1, 2,\ldots, 6$ samples per class. We did 200 random trials of training with $1, 2, \ldots, 6$  samples per class. These independent trials of training are tested on the remaining data. We then computed the average macro F1-score of the test data. Figure~\ref{Fig_F1_low_training_sample_regime_multi_class_classification} and Figure~\ref{Fig_standard_deviation_low_training_sample_regime_multi_class_classification} represents the average macro F1-score and standard deviation of macro F1-scores of test data for 200 independent trials of training with $1, 2, \ldots, 6$ samples per class respectively.

%
%
In the low training sample regime ChaosFEX+SVM slightly outperforms SVM with linear kernel with $1, 2, \ldots, 6$ training samples per class. With just one training sample ChaosFEX gave an average macro F1-score > 0.90. There is a consistent increase in the performance of ChaosFEX for the multiclass classification problem in the low training sample regime. The 200 random trials of training ensures that this high performance is not due to overfitting.
\subsection{SARS-CoV-2 vs. SARS-CoV-1}
The dataset corresponding to the classification of SARS-CoV-2 and SARS-CoV-1 are provided in Table~\ref{Table_SARS-CoV-2_vs_SARS-CoV-1}. The number of genome sequences of SARS-CoV-2 is higher when compared to the genome sequence of SARS-CoV-2. The maximum length of the genome sequence is 30129.  We did a five fold stratified crossvalidation on this data. The train and validation split (train (\%), Val (\%)) for  fold 1, fold 2, fold 3, fold 4 and fold 5 were chosen to be approximately ($80\%$, $20\%$). With the same  hyperparameters, ChaosFEX ($q = 0.34$, $b = 0.499$, $\epsilon = 0.183$) gave high performance in all the five folds. The results for the five fold validation for ChaosFEX and SVM with linear kernel (without ChaosFEX features) are provided in Table~\ref{Table_larger_data_binary_classification_five_fold_validation_neurochaos} and Table~\ref{Table_multi_class_classification_five_fold_validation_svm}.
\begin{table*}[!h]
\centering
\caption{\label{Table_larger_data_binary_classification_five_fold_validation_neurochaos} SARS-CoV-2 vs. SARS-CoV-1: Five fold validation for larger binary classification data using ChaosFEX+SVM.}
\begin{tabular}{|c|c|c|c|c|c|c|c|c|c|c|c|c|c|c|c|}
\hline
Folds & \multicolumn{3}{c|}{Fold - 1} & \multicolumn{3}{c|}{Fold - 2} & \multicolumn{3}{c|}{Fold - 3} & \multicolumn{3}{c|}{Fold - 4} & \multicolumn{3}{c|}{Fold - 5} \\ \hline
Metric & Pr & Re & F1 & Pr & Re & F1 & Pr & Re & F1 & Pr & Re & F1 & Pr & Re & F1 \\ \hline
Class-0 & 1 & 1 & 1 & 1 & 1 & 1 & 1 & 1 & 1 & 1 & 1 & 1 & 1 & 1 & 1 \\ \hline
Class-1 & 1 & 1 & 1 & 1 & 1 & 1 & 1 & 1 & 1 & 1 & 1 & 1 & 1 & 1 & 1 \\ \hline
\begin{tabular}[c]{@{}c@{}}Macro \\ Averaged\end{tabular} & 1 & 1 & 1 & 1 & 1 & 1 & 1 & 1 & 1 & 1 & 1 & 1 & 1 & 1 & 1 \\ \hline
\end{tabular}
\end{table*}

\begin{table*}[!h]
\centering
\caption{\label{Table_larger_data_binary_classification_five_fold_validation_svm} SARS-CoV-2 vs. SARS-CoV-1: Five fold crossvalidation for larger binary classification data using SVM with linear kernel (without ChaosFEX features).}
\begin{tabular}{|c|c|c|c|c|c|c|c|c|c|c|c|c|c|c|c|}
\hline
Folds & \multicolumn{3}{c|}{Fold - 1} & \multicolumn{3}{c|}{Fold - 2} & \multicolumn{3}{c|}{Fold - 3} & \multicolumn{3}{c|}{Fold - 4} & \multicolumn{3}{c|}{Fold - 5} \\ \hline
Metric & Pr & Re & F1 & Pr & Re & F1 & Pr & Re & F1 & Pr & Re & F1 & Pr & Re & F1 \\ \hline
Class-0 & 1 & 1 & 1 & 1 & 1 & 1 & 1 & 1 & 1 & 1 & 1 & 1 & 1 & 1 & 1 \\ \hline
Class-1 & 1 & 1 & 1 & 1 & 1 & 1 & 1 & 1 & 1 & 1 & 1 & 1 & 1 & 0.85 & 0.92 \\ \hline
\begin{tabular}[c]{@{}c@{}}Macro \\ Averaged\end{tabular} & 1 & 1 & 1 & 1 & 1 & 1 & 1 & 1 & 1 & 1 & 1 & 1 & 1 & 0.93 & 0.96 \\ \hline
\end{tabular}
\end{table*}
\begin{figure*}
\centering
	\begin{subfigure}{0.45\linewidth}
		\centering
		\includegraphics[width=1\linewidth]{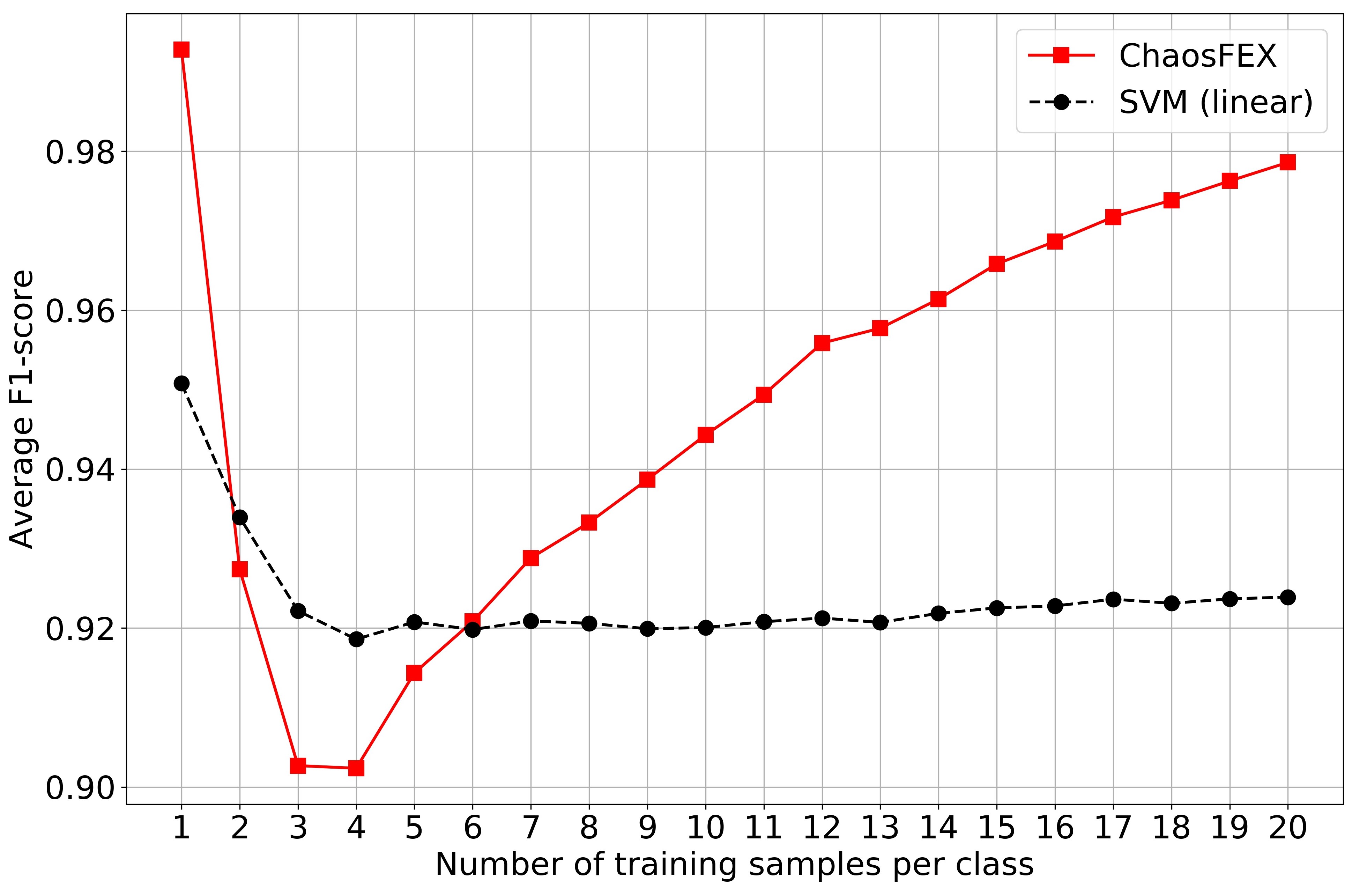}
		\caption{}\label{Fig_F1_low_training_sample_regime_larger_binary_classification}
	\end{subfigure}
	\begin{subfigure}{0.45\linewidth}
		\centering
		\includegraphics[width=1\linewidth]{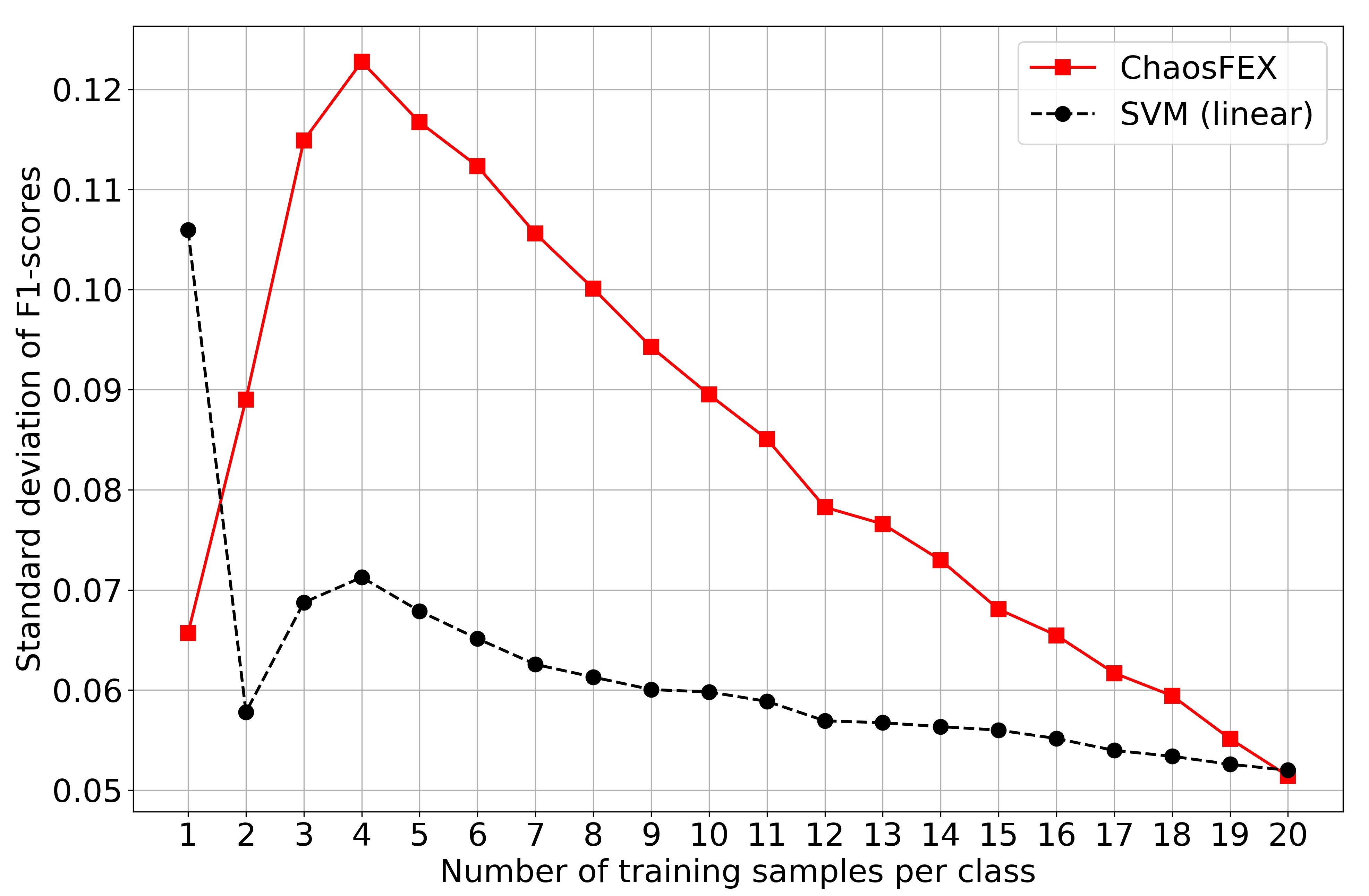}
		\caption{}\label{Fig_standard_deviation_low_training_sample_regime_larger_binary_classification}
	\end{subfigure}
\caption{SARS-CoV-2 vs. SARS-CoV-1 low training sample regime. (\subref{Fig_F1_low_training_sample_regime_larger_binary_classification}) Average macro F1-score of test data for 1000 random trials of training with $1, 2, \ldots, 20$ samples per class. (\subref{Fig_standard_deviation_low_training_sample_regime_larger_binary_classification}) Standard deviation of macro F1-scores for 1000 random trials of training with $1, 2, \ldots, 20$ samples per class.}
\label{fig:3:2}
\end{figure*}%
\subsubsection{SARS-CoV-2 vs. SARS-CoV-1: Low Training Sample Regime}
In the low training sample regime we used $1, 2,\ldots, 20$ samples per class. We did $1000$ random trials of training with $1, 2, \ldots, 20$  samples per class. These independent trials of training are tested on the remaining data.  Figure~\ref{Fig_F1_low_training_sample_regime_larger_binary_classification}  and Figure~\ref{Fig_standard_deviation_low_training_sample_regime_larger_binary_classification} represents the average macro F1-score and the standard deviation of macro F1-scores of testdata for $1000$ independent trials of training with $1, 2, \ldots, 20$ samples per class respectively.

In the case of low training sample regime for SARS-CoV-2 vs. SARS-CoV-1, we observe a maximum average macro F1-score $> 0.99$ for training with one sample per class. As the number of training samples increases, the average F1-score shows a decreasingly increasing trend. The standard deviation of F1-scores as number of training samples increases shows an increasingly decreasing trend. ChaosFEX slightly outperforms SVM with linear kernel in the low training sample regime except for training with 2, 3, 4 and 5 samples per class. Also in the five fold cross validation for the same data (Table ~\ref{Table_larger_data_binary_classification_five_fold_validation_neurochaos}) we get an average macro F1-score of 1.0 using ChaosFEX. The low training sample regime highlights the requirement of only a single sample of SARS-CoV-2 and SARS-CoV-1 for classification for ChaosFEX. From~\cite{compare_cov_2_cov_1}, SARS-CoV-2 and SARS-CoV-1 are genetically close to each other even though the SARS-CoV-2 is not a genetic descendent of SARS-CoV-1. However our experiments seems to indicate that the difference between the ChaosFEX features of the genomic sequences of the 2 viruses are significant enough that from very few observed sequences (few shot learning) ChaosFEX is able to generalize for efficient classification of larger sets of sequences from the 2 viruses.
\section{Conclusions\label{Section_conclusion}}
The combination of chaos and machine learning opens the possibility of developing brain inspired learning algorithms. In this work, we propose a \emph{Neurochaos Learning} (NL) architecture which explicitly employs chaotic neurons (unlike traditional ANNs which has simple dumb neurons) and by combining chaos-based feature extraction with SVM-based classification, we demonstrate efficacy and robustness of such an approach. Our proof of the Universal Approximation Theorem (UAT) is enabled by two properties of chaos - topological transitivity and existence of a dense orbit. An important benefit of our proof is the explicit construction of NL with the exact number of neurons needed to approximate a discrete time real valued function with finite support to any desired accuracy. Such an equivalent is not available for ANNs to the best of our knowledge. Thus the benefit of using the rich features of chaos is evident in our work.

In the experiments, we evaluated the performance of ChaosFEX both in low as well as high training sample regime for synthetically generated overlapping concentric circle data and coronavirus genome sequence data. In the case of classification of SARS-CoV-2 vs. SARS-CoV-1, ChaosFEX gave an average F1-score > 0.99 with just one training sample per class. This shows the robustness of the ChaosFEX features and its ability to generalize with very few training samples. The ChaosFEX features can be combined with any machine learning algorithm. The hyperparameters used in ChaosFEX preserves the performance of earlier learned tasks when more classes are added. This shows the efficacy of the proposed method to catastrophic forgetting problem. Combing ChaosFEX with Deep learning and Reinforcement learning algorithms are a future line of work.

The code used for the classification of OCCD can be found here: \url{https://github.com/HarikrishnanNB/occd_experiments}. The code used for classification of coronavirus genome sequence is available here:~\url{https://github.com/HarikrishnanNB/genome_classification}.

Accession IDs as well as acknowledgement of the genome sequence used in the classification of SARS-CoV-2 vs. SARS-CoV-1 is available in the GitHub repository:~\url{https://github.com/HarikrishnanNB/genome_classification/tree/master/sequence_usage_acknowledgements}.

`\section*{Acknowledgment}
Harikrishnan N. B. thanks ``The University of Trans-Disciplinary Health Sciences and Technology (TDU)'' for permitting this research as part of the PhD programme. The authors gratefully acknowledge the financial support of Tata Trusts. 
%
%
%
%
\bibliographystyle{unsrt}

\end{document}